\documentclass{article}
\usepackage{amssymb}
\usepackage{amsmath}

\usepackage{pifont}
\usepackage{enumitem}
\usepackage[table]{xcolor}
\usepackage{colortbl}
\usepackage{wrapfig}

\usepackage{amsthm}
\newtheorem{theorem}{Theorem}

\theoremstyle{plain}

\usepackage{multirow}
\usepackage{algorithm,algpseudocode}
\usepackage{graphicx}
\usepackage{caption}
\usepackage{subcaption}
\usepackage{float}

\definecolor{darkgreen}{rgb}{0.0, 0.6, 0.0} 

\def\dd{\mathrm{d}}

\usepackage[final]{corl_2025} 

\title{Train-Once Plan-Anywhere \\
Kinodynamic Motion Planning via Diffusion Trees}

\author{
  Yaniv Hassidof\thanks{Corresponding author; Email: yaniv\_hass@campus.technion.ac.il, tomj@campus.technion.ac.il, kirilsol@technion.ac.il.} \quad
  Tom Jurgenson 
  \quad  
  Kiril Solovey 
  \\
  Viterbi Faculty of Electrical and Computer Engineering \\ 
  Technion--Israel Institute of Technology, 
  Haifa, Israel \\
 }

\begin{document}
\maketitle


\vspace{-30px}
\begin{abstract}
Kinodynamic motion planning 
is concerned with computing collision-free trajectories while abiding by the robot's dynamic constraints. This critical problem is often tackled using sampling-based planners (SBPs) that explore the robot's high-dimensional state space by constructing a search tree via 
action propagations. Although SBPs can offer global guarantees on completeness and solution quality, their performance is often hindered by slow exploration due to uninformed action sampling. 
Learning-based approaches 
can yield significantly faster runtimes, 
yet they fail to generalize to out-of-distribution (OOD) scenarios and lack critical guarantees, e.g., safety, thus limiting their deployment on physical robots. 
We present Diffusion Tree (DiTree): a \emph{provably-generalizable} framework leveraging diffusion policies (DPs) as informed samplers to efficiently guide state-space search within SBPs.
DiTree combines DP's ability to model
complex distributions of expert trajectories, conditioned on local observations,
with the completeness of SBPs to yield \emph{provably-safe} solutions within a few action propagation iterations for complex dynamical systems. 
We demonstrate DiTree's power with an implementation combining the popular RRT planner with a DP action sampler trained on a \emph{single environment}.  
In comprehensive evaluations on OOD scenarios, DiTree achieves on average a 30\% higher success rate compared to standalone DP or SBPs, on both a dynamic car and Mujoco’s ant robot settings (for the latter, SBPs fail completely). Beyond simulation, real-world car experiments confirm DiTree’s applicability, demonstrating superior trajectory quality and robustness even under severe sim-to-real gaps.


Project webpage: 
\href{https://sites.google.com/view/ditree}{\underline{sites.google.com/view/ditree}.} 
\end{abstract}



\begin{figure}[H]
\vspace{-5pt}
    \centering
    \includegraphics[width=\linewidth, trim={0 60pt 0 20pt},clip]{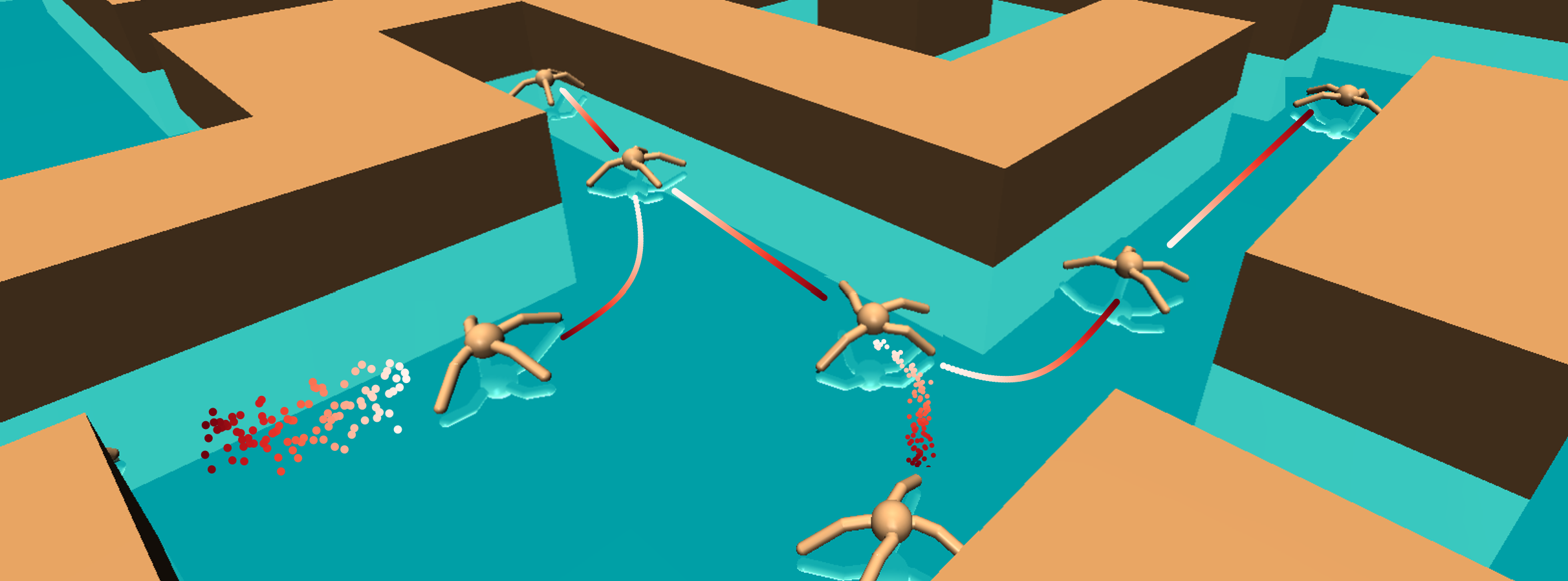}
        \caption{Visualization of DiTree on D4RL's AntMaze setting in a random maze \textit{unseen} during training. Edges are generated via diffusion.} 
    \label{fig:header}
    \vspace{-10pt}
\end{figure}

\vspace{-3mm}
\section{Introduction}    

\vspace{-2mm}
    Robotic mobility has been a longstanding goal in artificial intelligence, pivotal for real-world applications such as autonomous driving~\citep{AVsurvey}, drones~\citep{UAVsurvey}, and humanoid robots~\citep{Humanoid}. Such systems are often underactuated ~\citep{underactuated}, requiring motion planners to account not only for environmental constraints but also for the physical limitations and dynamics of the robot itself, in a problem called \emph{kinodynamic motion planning} (KMP)~\citep{lavalle2006planning}. 
    Despite its importance, KMP remains challenging due to the 
    complexity of searching high-dimensional state and action spaces with non-linear dynamics. 
    Echoing Sutton's~\cite{sutton2019bitter} “Bitter Lesson”, formidable challenges in Computer Science are often solved by large-scale general-purpose methods rather than specialized heuristics, with \textit{search} and \textit{learning} emerging as two approaches capable of arbitrary scaling, rendering them essential in tackling KMP.
    
    A cornerstone of search approaches for KMP are sampling-based planners (SBPs)~\citep{SBMP-review}. 
    By sampling random actions and simulating them through the dynamic model, SBPs grow a collision-free tree that spans the state space.
    SBPs are valued for their simplicity, versatility, and strong theoretical foundations. However, their exhaustive exploration can be inefficient in high-dimensional spaces, particularly in environments densely cluttered with obstacles or involving complex dynamics.


    Learning-based approaches have emerged as a promising alternative for improving motion planning efficiency. Prominent among them, diffusion planning~\citep{diffuser} casts planning as a form of probabilistic inference~\citep{controlasinference} and samples from the multi-modal distribution of expert behaviors to synthesize diverse and high-quality trajectories. Unfortunately, such models struggle to generalize to unseen environments, and inherently lack guarantees such as collision avoidance.
    

    \textbf{Contribution.} 
    Motivated by breakthroughs combining learning and search, e.g., AlphaGo~\citep{alphago,alphazero} and LLM reasoning~\citep{openai2024o1,deepseekr1}, we propose a learning-meets-search strategy for KMP.
    We introduce \emph{Diffusion Tree (DiTree)}, a novel framework that combines diffusion models (DMs) with sampling-based tree search for efficient KMP. DiTree leverages DMs as powerful, environment-aware motion priors to guide exploration toward promising trajectories. By integrating these priors with SBPs, DiTree ensures collision avoidance and dynamic feasibility, while 
    successfully navigating unseen environments. 
    We demonstrate that DiTree achieves substantial efficiency improvements and outperforms state-of-the-art methods across diverse tasks and robotic platforms. We explore the set of unique challenges and trade-offs entailed with employing a DM for tree search in our ablation studies.
    

    \textbf{Organization.} 
    We first survey related work (Sec.~\ref{sec:related_work}), provide background on KMP, SBPs, and DMs (Sec.~\ref{sec:preliminaries}), and then introduce our DiTree framework (Sec.~\ref{sec:Diffusion_Sampling_based_Planning}). 
    We then present implementation details and practical considerations, followed by a comprehensive evaluation (Sec.~\ref{sec:experiments}). 
    Sec.~\ref{sec:conclusion} concludes with a discussion and 
    future work, followed by limitations in Sec.~\ref{sec:limitations}.

\vspace{-4.5mm}
\section{Related Work}
\label{sec:related_work}
\vspace{-4mm}


    \textbf{Sampling-based planners}~\citep{lavalle2006planning} fall broadly into two categories: \emph{geometric} and \emph{kinodynamic} planners. Geometric SBPs~\citep{PRM,LazyPRM,PRMstar} construct a graph by sampling states in the configuration space and connecting them with straight edges. While conceptually simple and widely used, they assume the robot can move directly between any two states without regard for its dynamics. 
    \emph{Kinodynamic} SBPs, by contrast, incorporate the system’s differential constraints, such as bounded turning radius, velocity, or torque. To connect two states while respecting these constraints, some planners~\citep{kinoRRT*} employ a steering function~\citep{lavalle2006planning,BVP}—which computes a dynamically feasible trajectory between two configurations. However, exact steering functions are only available for a limited class of systems (e.g., linear model or a Dubins car).

    In contrast,tree-based kinodynamic SBPs~\citep{kinoRRT,SST,statecostspaceRRT}
    span a search tree by extending its branches by simulating random actions, thus ensuring the generated edges adhere to the dynamics. 
    Modern approaches attempt to enhance efficiency by incorporating heuristics and pruning~\citep{SST,KPIECE,statecostspaceRRT} while ensuring (asymptotic) optimality~\citep{AOAnalysis,statecostspaceRRT,IJRR}, meaning they converge to an optimal solution as the number of samples approaches infinity.
    While these methods allow scalable search for trajectories, their exhaustive exploration can be inefficient in large spaces or systems with high degrees of freedom. 

    \textbf{Learning-based approaches}~\citep{dataDriven} 
    leverage past experience 
    to improve solution quality and reduce computational effort when solving new planning problems.
    The seminal work~\cite{diffuser} 
    along with numerous follow-up studies~\citep{diffusionpolicy,EDMP,mpd_motionplanningdiffusion,safediffuser,potentialdiffusion}, explore the use of diffusion models (DMs)~\citep{diffusion,ddpm,song2019generative} as well as the recent Flow Matching (FM)~\citep{lipmanflowm,rectifiedflow}.
    DMs have demonstrated strong capabilities in planning, particularly due to their ability to sample from high-dimensional, multi-modal trajectory distributions. However, we identify several critical limitations that remain widespread: 
    \begin{itemize}[leftmargin=*]
        \item \textbf{Lack of Collision Avoidance Guarantees}. Trajectory prediction via DMs could result in collisions with the environment or  the robot itself. Although custom controllers for 
        collision avoidance during physical manipulator 
        experiments can be employed~\cite{diffusionpolicy}, they are difficult to craft for agile fast-moving robots. Other methods~\citep{EDMP,mpd_motionplanningdiffusion,PRESTO,potentialdiffusion,refine_diffusion,DiffusionIn3D,feng2024resisting}  mitigate the issue by guiding the denoising process away from obstacles, but they do not guarantee the final output to be completely collision-free, even after post processing refinement. 
        In contrast, the incorporation of collision checking in DiTree means our search tree is inherently collision-free. 
        
        \item \textbf{Limited Out-of-Distribution (OOD) Performance}. As is endemic to many learning-based approaches, DMs tend to catastrophically fail when confronted with environments that diverge sharply from the training set. 
        Some methods~\citep{simpleHierarchical,litediffuser} attempt to alleviate this issue through hierarchical planning, yet 
        generalization is largely confined to environments resembling the training set, with the work~\citep{mpd_motionplanningdiffusion} explicitly opting to specialize in a single environment due to the impractical data demands of broader generalization.
        In comparison, We empirically demonstrated that for DiTree training on a single maze is sufficient to generalize to wildly different environments. 

        \item \textbf{Predefined Trajectory Horizon}. DMs operate by denoising a fixed-length input, requiring the trajectory horizon to be specified \textit{a priori}. A common workaround is to use the DM as a Model Predictive Control (MPC) policy~\citep{diffusionpolicy}, replanning iteratively over a short horizon.
        Other approaches~\citep{EDMP,mpd_motionplanningdiffusion,dipper,simpleHierarchical}  select a predefined heuristic horizon, either committing to it or interpolating between predicted states but losing their temporal significance. 
        In contrast, DiTree's tree-based search dynamically expands edges until the goal is reached, sidestepping these limitations.

        \item \textbf{Lack of Differential Constraints}. 
        Many diffusion-based 
        approaches~\citep{diffuser,dipper,safediffuser} directly predict trajectory states without explicitly enforcing differential constraints in the final outputs, thus requiring an external local planner to avoid violation of the robots kinodynamic constraints~\citep{dipper}.
        In contrast, our approach predicts actions and applies them directly to the robot's dynamic model, ensuring inherent adherence to system dynamics.
 
    \end{itemize}

    \textbf{Hybrid approaches} have also explored the integration of learning and search. While AlphaZero~\citep{alphazero} is a notable example, its reliance on MCTS impedes its use in continuous-action KMP, and its learned value function restricts its generalization to new environments. Most other works focus on motion planning without differential constraints, or address higher-level task planning~\citep{qureshi2019motion,Ichter2017,NeuralRRTstar,NeuralEETrees,BroadlyExploringLT}. Existing methods which do address KMP typically combine a low-level learned policy with a high-level tree search, but provide the policy with either no obstacle information~\citep{ImprovingKP,TerrainAware} or only limited observation from onboard sensors~\citep{RLRRT}. Notably, RL-RRT~\citep{RLRRT} acknowledge reliance on simulated LIDAR measurements alone hampers planning in complex maze-like environments, especially those unseen during training (see their Fig.~7).
    In contrast, our diffusion model is privy to the entire local map regardless of visibility. 
    Furthermore, to the best of our knowledge, none of these methods address planning challenges for robots characterized by large state or action spaces. 

\vspace{-2mm}
\section{Preliminaries and Problem Definition}
\label{sec:preliminaries}

\vspace{-2mm}

\textbf{Kinodynamic Motion Planning.} Consider a robot operating within a continuous state space $\mathcal{X} \subseteq \mathbb{R}^n$ and a control space $\mathcal{U} \subseteq \mathbb{R}^m$, governed by the dynamics $\dot{x} = f(x, u),  x \in \mathcal{X}, u \in \mathcal{U}$,
where \( f: \mathcal{X} \times \mathcal{U} \to \mathbb{R}^n \). 
Denote by $\mathcal{X}_{\text{free}} \subset \mathcal{X}$  the set of collision-free states.
For a start state \( {x}_\text{start} \) and goal region \( \mathcal{X}_\text{goal} \subset  \mathcal{X} \), the objective 
is to compute a control function \( \mathbf{u}: [0, T] \to \mathcal{U} \), for some $T>0$, that induces a 
trajectory \( \tau: [0, T] \to \mathcal{X}\) satisfying $\tau(0) = x_\text{start},\;\tau(T) \in \mathcal{X}_\text{goal},\;\dot{\tau}(t)=f(\tau(t),\mathbf{u}(t)),\;\tau(t)\in\mathcal{X}_{\text{free}},\;\forall t\in[0,T].$
I.e.,  the trajectory must remain collision-free, 
respect the dynamics, and reach the goal region. 
Importantly, 
there is generally no closed-form solution to this problem. Consequently, sampling-based algorithms are widely used to obtain feasible trajectories.


\algtext*{EndFor}
\algtext*{EndIf}
\begin{wrapfigure}[11]{R}{0.50\textwidth} 
\vspace{-16pt} 
\begin{minipage}{0.49\textwidth}
\begin{algorithm}[H]
    \caption{Sampling-based Tree Planner}
    \label{alg:SBP}
    \begin{algorithmic}[1]
        \State $\mathcal{T}.\text{init}(x_{\text{start}})$
        \For{$i = 1$ to $k$}
            \State $x_{\text{near}} \gets \text{Node\_Selector}(\mathcal{T})$
            \State $u \gets \text{Action\_Selector}(\mathcal{U})$
            \State $(\pi,x_{\text{new}}) \gets \text{Propagate}(x_{\text{near}}, u)$
            \If{$\text{Collision\_Free}(x_{\text{near}}, x_{\text{new}}, \mathcal{X_\text{obs}})$}
                \State $\mathcal{T}.\text{add\_vertex}(x_{\text{new}})$
                \State $\mathcal{T}.\text{add\_edge}(x_{\text{near}}, x_{\text{new}},\pi)$
            \EndIf
        \EndFor
        \State \textbf{return} $\mathcal{T}$
    \end{algorithmic}
\end{algorithm}
\end{minipage}
\end{wrapfigure}

     \textbf{Sampling-based Planning.} Tree-based SBPs~\citep{RRTconnect,kinoRRT,SST} sample states and actions to generate a dynamically-feasible tree. 
     By construction, these algorithms ensure that any returned solution adheres to both collision-avoidance and kinodynamic constraints. We focus on this family of approaches and, unless stated otherwise, use the term SBP to refer specifically to tree-based methods. We refine this notion into a general blueprint presented in Alg.~\ref{alg:SBP}. 
     Rather than depicting a specific algorithm, this outline captures the fundamental logic that underpins most kinodynamic SBPs. 
     Our framework is applicable to any method built around this core logic, e.g.,~\citep{kinoRRT,statecostspaceRRT,SST,IJRR}.

    The SBP outline begins by initializing a tree with the start state ${x}_\text{start}$ (Line~1). We then proceed to iteratively select some tree node \(x_{near}\) (Line~3),  a control input (Line~4), and employ forward integration of the dynamic model to simulate the control, termed forward propagation (Line~5), resulting in a subtrajectory $\pi$ from $x_{near}$ to $x_{new}$. If the resulting branch is collision free it is added to the tree. For instance, in  (Kinodynamic-)RRT, node selection (Line~3) is implemented by sampling a random state \( x_{\text{rand}} \in \mathcal{X} \) and identifying the nearest existing node \( x_{\text{near}} \) in the tree based on a predefined distance metric, after which action selection (Line~4) is done using uniform random sampling. 

\textbf{Diffusion and Flow Matching Policies.} Diffusion~\citep{diffusion,ddpm,song2019generative} and flow matching~\citep{lipmanflowm,rectifiedflow} models are generative frameworks designed to efficiently sample from complex distributions by learning a mapping $f_{\theta}: R^d \rightarrow R^d$ that transforms a noise sample \(x_0 \sim \mathcal{N}(0,I)\) to a sample from the target distribution \(x_1 \sim D\). 
Both models 
 excel at generating complex high-dimensional data such as images~\citep{SD3}, protein backbones~\citep{protein}, and robotic trajectories~\citep{diffuser,diffusionpolicy}.
\emph{Diffusion models} (DMs) learn to reverse a gradual noising process, where Gaussian noise is progressively added to a data point \(x_0\) until it becomes pure noise \(x_1\). By training a neural network to iteratively denoise \(x_1\), it can reconstruct realistic samples from noise.
\emph{Flow matching models} (FMs) learn 
a velocity field
\( v_\theta(x, t) \) that describes the optimal direction of movement at each point: $\frac{dx}{dt} = v_\theta(x, t)$.
By integrating this velocity field, samples are transported from an initial noise sample~\( x_0 \) to a final structured data point~\( x_1 \). 
FM achieves superior performance when using fewer integration steps~\citep{SD3,AdaFlow}.



Both DMs and FMs can serve as robotic policies, referred to as Diffusion Policies~\citep{diffusionpolicy}, which generate sequences of robotic actions or future states $u_{1:N}$ by sampling from a conditional distribution 
\( p(u|x, g, o) \). 
These models are often conditioned on the robot’s current state \( x \), the desired goal state \( g \), and sensor observation \( o \) such as RGB images, depth maps~\citep{diffusionpolicy}, or point clouds~\citep{3Ddiffusion}. 


\vspace{-3mm}
\section{Diffusion Tree Algorithm}
\label{sec:Diffusion_Sampling_based_Planning}
\vspace{-3.1mm}

We present our DiTree approach for leveraging diffusion models (DMs) for efficient kinodynamic motion planning (KMP) and discuss its theoretical implications.  

\setlength{\belowcaptionskip}{-5pt} 


\begin{figure}[!htbp]
    \centering
    \includegraphics[width=0.93\linewidth]{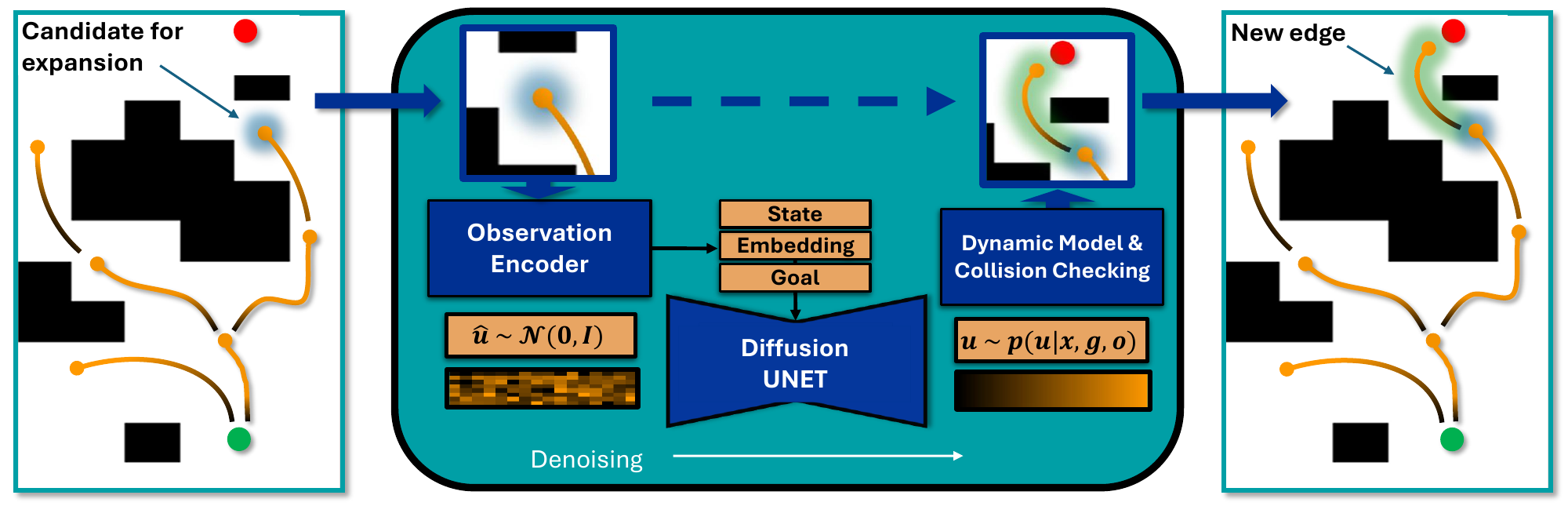}
    \caption{
    Action sampling in DiTree:  
    (Left) A candidate node is selected for expansion and  a local observation is extracted from its surroundings.  
   (Center) A DM conditioned on the observation, current robot state, and goal, generates an action sequence which is simulated and checked for collision. 
    (Right) The new edge is added to the tree. Start indicated by \textcolor{darkgreen}{\ding{108}}, goal by \textcolor{red}{\ding{108}}.
    }
    \label{fig:method}
    \vspace{-2mm}
\end{figure}

\vspace{-2.3mm}
\subsection{Action Sampling with Diffusion Policy}
\label{subsection:action_sampling}
\vspace{-2mm}



Our framework enhances 
methods compatible with 
Alg.~\ref{alg:SBP}, by implementing its action selection procedure (Line~4) using a context-aware sampling procedure. 
Instead of sampling a single action from a uniform distribution, we sample a sequence of $N$ actions $u_{1:N}$, drawn from a distribution informed by the planning context. 
Inspired by recent advances in diffusion-based planning~\citep{diffuser,diffusionpolicy}, we implement action sampling $u_{1:N} \sim p\bigl(u_{1:N} \mid x_{\text{near}},{x}_{\text{target}}, \mathcal{X}_{\text{obs}}^{\text{near}}\bigr)$ as inference from a \emph{conditional diffusion policy} (see Fig.~\ref{fig:method}). The policy is trained on expert trajectories and conditioned on three key inputs: (1) the current tree state \(x_{\text{near}}\), (2) a target state (which is either a goal or an exploration-guiding state) \(x_{\text{target}}\), and (3) local obstacle information \(\mathcal{X}_\text{obs}^{\text{near}}\) with respect to \(x_{\text{near}}\). 
During training, \(x_{\text{target}}\) is  set to the scenario’s goal state, reinforcing goal-directed behavior. However, at test time, we introduce a \emph{diffusion goal bias (DGB)} to promote exploration, replacing the goal state with a randomly sampled intermediate state \(x_{\text{rand}}\). This encourages the policy to 
explore alternative regions of the space, generating more diverse actions and improving search coverage. 

To foster generalization in our 
approach, we simplify the learning problem through two key strategies: (1)  
As effective planning should be agnostic to arbitrary global coordinates, we represent $x_{\text{target}}$ relative to the chosen state $x_{\text{near}}$,  introducing \emph{translation and rotation invariance}. 
(2) As training a sampler that effectively conditions on all obstacles in a scene across arbitrary scenarios requires a highly diverse and extensive dataset,  
we condition instead the DP on a \emph{localized, state-dependent subset} of obstacles, denoted as $\mathcal{X}_\text{obs}^\text{near}$, 
extracted relative to the frame of $x_\text{near}$.

After action sampling, a new edge for tree expansion is generated by executing the sequence \(u_{1:N}\) via forward propagation, where each action \(u_i\) is applied for some $dt$. 
The resulting trajectory segment is accepted only if it passes collision checking, otherwise it is discarded.
To ensure full information utilization during edge generation---despite relying only on local observations---we employ  DM as an MPC policy during forward propagation. I.e., 
the policy iteratively resamples actions based on the updated local observations, enabling a flexible generation of arbitrarily long edges by perpetual sampling and forward propagation even with a very local view. The process continues until a termination condition is met, such as reaching a predefined propagation duration. 

Action sampling with diffusion models is substantially more time-consuming than other components of Alg.~\ref{alg:SBP}, becoming the primary bottleneck in per-iteration runtime. This challenges their integration into sampling-based planners (SBPs), which require frequent sampling to drive exploration. Unlike standalone MPC-style diffusion policies~\citep{diffusionpolicy}, where the number of denoising steps is tuned to match a target control frequency, tree-based planners have no natural temporal anchor for setting the sampling schedule. Instead, the objective is to maximize planning success within a fixed time budget, requiring a careful balance between sampling speed and action quality. Fast sampling enables exploration of multiple branches, while not sacrificing the quality needed to capture the multimodal distribution of feasible edges. In Sec.~\ref{sec:experiments}, we empirically investigate this trade-off and demonstrate that flow matching offers an effective balance between efficiency and performance. Consequently, hyperparameters such as the number of denoising steps must be reconsidered for the unique demands of tree-based search.


\vspace{-1.5mm}
\subsection{Theoretical Guarantees}
\label{theoretical_guarantees}
\vspace{-3mm}


Some SBPs have guarantees such as \emph{probabilistic completeness} (PC)~\citep{RRTPC} and \emph{asymptotic optimality} (AO)~\citep{SST}, which  rely on the \emph{full support} of uniform random action sampling. Although DiTree uses non-uniform diffusion-based samples from a learned distributions, 
we demonstrate that the favorable SBP properties can be transferred to our approach, based on our proof that diffusion models posses full support. We prove the following result for RRT-style node selection (line~3). 



\begin{theorem}\label{thm:rrt_pc}
Consider a Lipschitz-continuous system $f: \mathcal{X} \times \mathcal{U} \to \mathbb{R}^n$, and suppose that the action sampler $u_{1:N} \sim p(u_{1:N} \mid \cdot )$ has full support. Then RRT-based DiTree is PC: there exist constants $a,b>0$ such that for any 
robust KMP problem, DiTree finds a solution with probability $\geq 1-ae^{-bk}$, where $k$ is the number of samples.  
\end{theorem}


The full proof, which relies on an intimate understanding of RRT's behavior, along with a derivation of full support for commonly-used DMs and FMs, are provided in Appendix~\ref{sec:proof}. We also discuss how asymptotic optimality is achieved when pairing DiTree with planners such as SST~\cite{SST} and AO-RRT~\cite{statecostspaceRRT}.

\vspace{-3.5mm}
\section{Experiments}
\label{sec:experiments}

\vspace{-3mm}

\begin{figure}[!htbp]
    \centering
    \vspace{-2mm}
    \includegraphics[width=\linewidth]{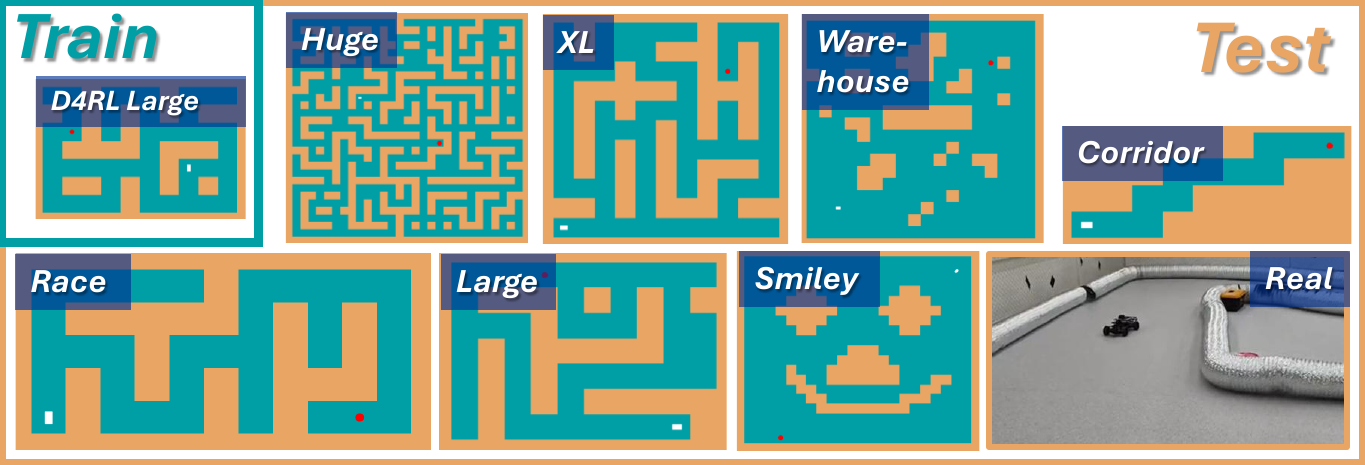}
    \caption{Experiment scenarios: We train on a single map, D4RL \textit{AntMaze Large} ({Top Left}), and test on a variety of unseen maps. Queries shown for reference, car in white and goal in \textcolor{red}{\ding{108}}.}
    \label{fig:maps}
\end{figure}

To assess generalization and performance, we test DiTree against several baselines across 15 distinct scenarios on unseen maps (Fig.~\ref{fig:maps}), using two robot types. Each method is run for 20 trials per scenario with a 120-second time limit. Performance is measured by the success rate (i.e., the proportion of trials that result in a collision-free path) and the average planning time over successful trials. 

We implement an instance of DiTree in Python. As an SBP backbone we build upon RRT~\citep{kinoRRT} 
due to its simplicity and popularity.
For the diffusion policy we adopt single-step flow matching (FM)~\citep{lipmanflowm}, due to a favorable trade-off between speed and quality compared to diffusion models~\citep{AdaFlow}. 
As a single planning query could require running inference hundreds of times, we prioritize inference speed over absolute quality, but explore this trade-off in our ablation study. 
While more elaborate methods for single-step inference exist~\cite{consistencyfm,Shortcut}, 
they introduce a more complex training pipeline.
For our local observations we use an occupancy grid (Fig~\ref{fig:method}).
Transformers are widely used and generally yield superior results, yet they are also highly sensitive to hyperparameters~\citep{diffusionpolicy,Dasari2024TheIF}. 
We leave their integration for future work and instead use UNet~\citep{UNET}.

\textbf{Robots.}
We evaluate our method on two robotic domains: our custom \textit{CarMaze} (implemented using the CasADi framework~\citep{casadi,carModel}) and D4RL's \textit{AntMaze} (part of the D4RL benchmark; based on the MuJoCo physics engine~\citep{mujoco, gym_robotics}), both operating within 2D maze layouts. CarMaze relies on a 6D non-holonomic ground vehicle modeled using a single-track dynamic system~\citep{carModel}, controlled via the throttle rate and steering rate. AntMaze consists of a 29D quadruped robot with a 8D action space specifying joint torques (Fig.~\ref{fig:header}). See  App.~\ref{sec:car_dynamics} for more details.
For each robot type, an FM policy is trained on an offline dataset collected from a \emph{single} environment, D4RL's~\citep{gym_robotics,Minari} \textit{Large} maze. 
To encourage relevant planning behaviors, we filter out states in close proximity to obstacles. 
Training and testing were conducted on an RTX 3090 GPU. Training lasts 2 hours.

\textbf{Baselines.} We compare DiTree against the two classical SBPs RRT and SST, which are implemented in OMPL~\citep{OMPL}---a leading C++ implementation. 
\begin{figure}
    \vspace{-5mm}
    \centering
    \includegraphics[width=0.95\linewidth]{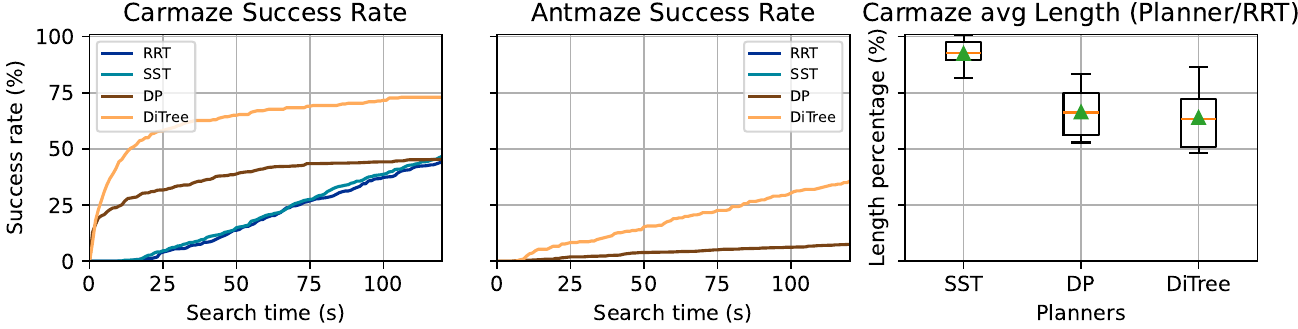}
    \caption{({Left and middle}) Average success rate vs. search time across all test scenarios for CarMaze and AntMaze, comparing different methods. ({Right}) Comparison of the distribution of average trajectory lengths across scenarios, relative to RRT. }
    \label{fig:SR_and_length}
    \vspace{-3mm}
\end{figure}
For a learning-based baseline, as far as we know there are currently no available implementations of kinodynamic methods that could be trained on a single environment and generalize to others. Therefore, we compare against a standalone DP~\citep{diffusionpolicy}, using the same model checkpoint as DiTree. This baseline highlights the difference between direct trajectory generation and action sampling in an SBP. For the DP setup, the policy employs the dynamic model for forward propagation of sampled action sequences, but implements search by repeatedly generating rollouts until either the goal is reached without collision or timing out. 

\vspace{-3mm}
\subsection{Results}
\label{subsec:Results}
\vspace{-2mm}

\begin{table}[h!]
    \centering
    \vspace{0.5mm}
    \scriptsize
    \resizebox{\linewidth}{!}{
    \begin{tabular}{|c|cccc|cc|}
        \hline
        \multirow{2}{*}{\textbf{Scenario}} 
          & \multicolumn{4}{c|}{\cellcolor[HTML]{009FA6}{\color{white}\textbf{CarMaze (SR\,(\%) $\uparrow$ / RT\,(s) $\downarrow$)}}} 
          & \multicolumn{2}{c|}{\cellcolor[HTML]{E8A563}\textbf{AntMaze (SR\,(\%) $\uparrow$ / RT\,(s) $\downarrow$)}} \\
        \cline{2-7}
          & \cellcolor[HTML]{009FA6}{\color{white}RRT} & \cellcolor[HTML]{009FA6}{\color{white}SST} & \cellcolor[HTML]{009FA6}{\color{white}DP} & \cellcolor[HTML]{009FA6}{\color{white}DiTree(ours)} 
          & \cellcolor[HTML]{E8A563}DP & \cellcolor[HTML]{E8A563}DiTree(ours) \\
        \hline
        \hline
Race          & 0 / -- & 0 / -- & 5 / 93.5 \( \pm \) 0.0 & \textbf{45} / \textbf{40.3} \( \pm \) 24.6 & 0 / -- & \textbf{30} / \textbf{87.7} \( \pm \) 18.4 \\
Warehouse 1   & 50 / 90.2 \( \pm \) 22.4 & 55 / 95.1 \( \pm \) 19.6 & 0 / -- & \textbf{95} / \textbf{21.7} \( \pm \) 15.5 & 0 / -- & \textbf{45} / \textbf{87.2} \( \pm \) 17.7 \\
Warehouse 2   & 20 / 91.2 \( \pm \) 23.1 & 65 / 98.4 \( \pm \) 17.4 & \textbf{100} / 23.2 \( \pm \) 19.8 & \textbf{100} / \textbf{9.3} \( \pm \) 7.5 & 0 / -- & \textbf{25} / \textbf{94.3} \( \pm \) 14.0 \\
Warehouse 3   & 95 / 65.5 \( \pm \) 19.0 & 90 / 68.3 \( \pm \) 17.5 & 70 / 51.0 \( \pm \) 26.1 & \textbf{100} / \textbf{31.3} \( \pm \) 31.5 & 0 / -- & \textbf{10} / \textbf{106.3} \( \pm \) 6.4 \\
Corridor      & \textbf{100} / 66.8 \( \pm \) 20.0 & \textbf{100} / 57.1 \( \pm \) 24.4 & \textbf{100} / 24.2 \( \pm \) 20.1 & 10 / 44.2 \( \pm \) 23.0 & 60 / 47.4 \( \pm \) 39.8 & \textbf{100} / \textbf{19.1} \( \pm \) 19.8 \\
Huge 1        & 35 / 88.2 \( \pm \) 19.8 & \textbf{50} / \textbf{70.5} \( \pm \) 17.5 & 0 / -- & 5 / 115.2 \( \pm \) 0.0 & 0 / -- & \textbf{50} / \textbf{62.2} \( \pm \) 25.5 \\
Huge 2        & 50 / 58.6 \( \pm \) 31.6 & 50 / 56.7 \( \pm \) 28.5 & \textbf{100} / \textbf{0.5} \( \pm \) 0.4 & \textbf{100} / 4.0 \( \pm \) 5.7 & \textbf{35} / 66.0 \( \pm \) 32.2 & 20 / \textbf{20.3} \( \pm \) 13.6 \\
Huge 3        & 95 / 61.0 \( \pm \) 18.3 & 85 / 50.5 \( \pm \) 18.9 & 0 / -- & \textbf{100} / \textbf{14.0} \( \pm \) 16.0 & 5 / \textbf{19.6} \( \pm \) 0.0 & \textbf{80} / 51.6 \( \pm \) 31.7 \\
Large 1       & 65 / 91.2 \( \pm \) 14.3 & 50 / 95.6 \( \pm \) 16.7 & \textbf{100} / \textbf{2.1} \( \pm \) 1.7 & \textbf{100} / 19.9 \( \pm \) 27.1 & 10 / 74.4 \( \pm \) 31.5 & \textbf{50} / \textbf{66.6} \( \pm \) 27.8 \\
Large 2       & 0 / -- & 0 / -- & 5 / \textbf{12.9} \( \pm \) 0.0 & \textbf{45} / 41.7 \( \pm \) 29.9 & 0 / -- & \textbf{20} / \textbf{63.2} \( \pm \) 4.7 \\
XL 1          & 50 / 79.0 \( \pm \) 19.4 & 55 / 79.9 \( \pm \) 21.3 & 30 / \textbf{33.2} \( \pm \) 19.4 & \textbf{75} / 50.7 \( \pm \) 29.2 & 0 / -- & \textbf{60} / \textbf{66.2} \( \pm \) 22.9 \\
XL 2          & 0 / -- & 0 / -- & 0 / -- & \textbf{65} / \textbf{43.0} \( \pm \) 24.1 & \textbf{0} / -- & \textbf{0} / -- \\
XL 3          & 0 / -- & 0 / -- & 0 / -- & \textbf{60} / \textbf{51.0} \( \pm \) 26.9 & \textbf{0} / -- & \textbf{0} / -- \\
Smiley 1      & 15 / 95.4 \( \pm \) 21.9 & 10 / 71.7 \( \pm \) 32.1 & 80 / 51.4 \( \pm \) 30.2 & \textbf{100} / \textbf{8.6} \( \pm \) 9.5 & 0 / -- & \textbf{15} / \textbf{97.3} \( \pm \) 14.2 \\
Smiley 2      & \textbf{100} / 31.0 \( \pm \) 12.8 & \textbf{100} / 29.6 \( \pm \) 14.6 & \textbf{100} / \textbf{0.7} \( \pm \) 0.6 & \textbf{100} / 5.1 \( \pm \) 9.1 & 0 / -- & \textbf{30} / \textbf{74.9} \( \pm \) 22.7 \\
\hline
Average       & 45.0 / 67.3 \( \pm \) 27.5 & 47.3 / 66.2 \( \pm \) 29.6 & 46.0 / 20.7 \( \pm \) 26.9 & \textbf{73.3} / 23.3 \( \pm \) 26.9 & 7.3 / \textbf{54.5} \( \pm \) 38.0 & \textbf{35.7} / 59.4 \( \pm \) 34.0 \\
\hline

    \end{tabular}
    } 
    \vspace{1mm}
    \caption{Combined comparison for CarMaze (RRT, SST, DP, DiTree) and AntMaze (DP, DiTree) across different scenes. Each cell displays the ratio  success-rate / run-time (mean $\pm$ std), best result per scenario is \textbf{highlighted}. RRT and SST
    failed on all AntMaze trials.}
    \label{tab:combined_comparison}
    \vspace{-7mm}
\end{table}

Our results (Table~\ref{tab:combined_comparison}), show that DiTree achieves a strong performance across a range of out-of-distribution (OOD) scenarios, combining the speed of learned policies with the reliability of structured search. DiTree matches the runtime of DP--4× faster than classical SBPs---while  outperforming both in terms of success rate (on average). 
On CarMaze, DiTree achieves a 26\% higher success rate compared to all other methods, while in the AntMaze domain—where classical SBPs fail to produce any valid trajectories—it outperforms DP by a margin of 28\%.
While this work does not tackle optimal planning, a comparison of relative trajectory lengths (Fig~\ref{fig:SR_and_length}) shows DiTree discovers 25-50\% shorter trajectories compared to RRT, underscoring its strength in producing efficient solutions and showing promising potential for integration of asymptotically-optimal SBPs as the planning backbone.
These results demonstrate the advantage of integrating learned priors with guided exploration, particularly in environments characterized by complex dynamics and constrained, narrow passages.

There are a few CarMaze scenarios where DiTree underperforms. 
In the \textit{Corridor} scenario, for example, DiTree achieves only a 10\% success rate compared to 100\% for all other methods. 
Examining the search tree reveals that most of the tree nodes have reached configurations from which the car cannot possibly navigate towards the goal without collision. Upon sampling of such nodes, precious iterations are wasted in such local traps.  
This inefficiency could potentially be mitigated by 
incorporating pruning and advanced node selection strategies.

\vspace{-1.5mm}
\subsection{Ablation Study}
\label{sec:ablation}
\vspace{-0.1mm}



\begin{figure}[!htbp]
    \centering
    \vspace{-4.4mm}
    \includegraphics[width=\linewidth]{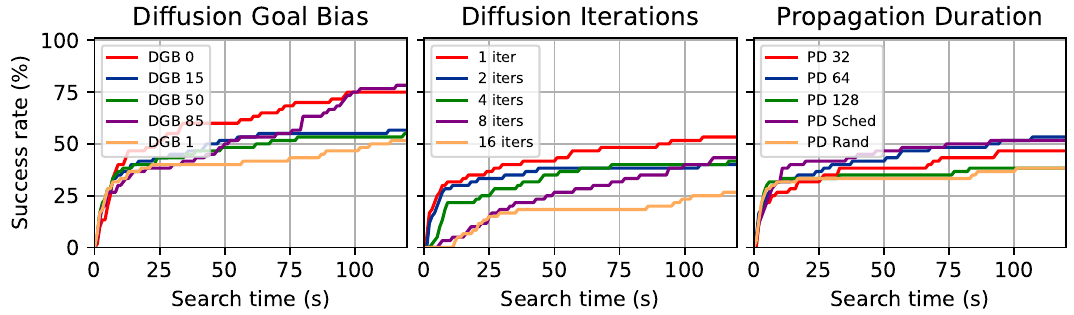}
    
    \caption{Comparison of Success Rate vs Runtime for different ablation tests. }
    \label{fig:ablation}
    \vspace{-2mm}
\end{figure}



We conduct an ablation study using our validation scenarios to assess the impact of key design choices specific to our framework. Results are presented in Fig~\ref{fig:ablation}. 
First, we vary the number of \emph{denoising iterations} used to generate a sample with FM model, exploring the trade-off between sampling speed and action quality. Surprisingly, we find a \emph{single diffusion iteration} yields the best overall performance, which we subsequently chose for our main experiments.
This stands in contrast to traditional diffusion-based planners, where even SOTA FM policies~\citep{AdaFlow} require several iterations to produce high-quality actions necessary for successful rollouts. In DiTree, however, faster, coarser samples are preferable, as they provide timely guidance for tree expansion without incurring the computational cost of multiple denoising steps. These results highlight a key distinction in how generative models are used within our framework: not to execute entire trajectories, but to prioritize promising search directions.

Next, we ablate the effect of a key hyperparameter of DiTree, \emph{propagation duration} ($N$). 
The propagation duration is defined as the length of the action sequence forming a tree edge. 
A shorter duration limits per-edge advancement—quickly shifting focus between nodes to broaden overall exploration—whereas a longer duration can cover more ground but risk inefficiency if a collision occurs. 
We compare several strategies for setting $N$: (i) a random selection between 32–128 steps per iteration, (ii) fixed values of 32, 64, or 128 steps, and (iii) a scheduled approach that gradually shortens edge length $(128\rightarrow96\rightarrow64)$ upon repeated node visits. Our experiments show a fixed value of 64 gave the best average results across our scenarios by a slight margin. 


Finally, we examine the role of \emph{diffusion goal bias} (DGB)\footnote{Not to be confused with RRT's node selection goal bias.}---a  hyperparameter controlling the ratio of conditioning the FM on $x_\text{rand}$ versus the final goal $x_{\text{goal}}$. We find that setting DGB to either 0\% or 85\% yields the highest overall success rates. We observe complex scenarios benefit from a DGB of 0\% due to the extensive exploration, however, simpler scenarios fall short as excessive random exploration leads to markedly longer and less efficient trajectories. 
Thus, in our experiments we opt for 85\% 
promoting rapid expansion toward the goal while maintaining sufficient exploration.

\vspace{-3.5mm}
\subsection{Real-World Applicability}
\label{sec:real_world}
\vspace{-1.5mm}
We demonstrate integration with a physical robot by comparing DiTree to RRT in a sharp-corner turning car scenario (Fig~\ref{fig:maps}). For deployment, output trajectories are smoothed and tracked using a pure-pursuit controller, without further optimization. Each planner is tested over ten runs (Appendix~\ref{sec:full_results}).
During execution, RRT encounters eight collisions, while DiTree consistently steers clear of obstacles. As the underlying algorithm is identical for both methods, this highlights the robustness and quality of sampling learned from the expert dataset, whose trajectories inherently maintain a safety margin—mirroring the strong performance of diffusion-based approaches in driving tasks demonstrated by prior work~\citep{diffdrive}.

\vspace{-2mm}
\section{Conclusion}
\label{sec:conclusion}

\vspace{-2mm}
We introduced DiTree---a novel framework that integrates diffusion models with sampling-based tree search to address the challenges of KMP. By leveraging learned, environment-aware priors for action sampling, DiTree balances the strengths of generative models and classical planners—achieving fast, generalizable, and collision-free planning, while preserving guarantees of the underlying planner.  

We aim for this work to lay the groundwork for safer, more general, data-driven motion planners in robotics. Future work includes 
inference optimization using distillation and quantization~\citep{PruneAndQuant}. Other potential improvements involve learning other SBP components like node selection, as well as better leveraging GPU parallelization during search.
As our work diverts the computational bottleneck of SBPs from collision detection~\cite{KleinbortSH16} to action sampling via DM (Appendix~\ref{subsection:full_results}), 
it motivates the development of new tailored models that exploit the problem structure to facilitate faster solutions of multiple queries.


\clearpage
\section{Limitations}\label{sec:limitations}

  




Although DiTree demonstrates strong performance, several assumptions limit its ability to reach its full potential in more complex settings. 

First, we assume full knowledge of all obstacles at planning time, including their geometry. 
This restricts applicability in dynamic or partially observed settings. One remedy is to extend Alg.~\ref{alg:SBP} to support partial knowledge of the environment~\cite{ElhafsiIJP20,KoenigLikhachev05} and replan over short horizons using updated observations.

Second, in our implementation, we rely on an approximation of the dynamic model, which could lead to reduced performance in practice, especially for complex real-world systems with chaotic dynamics or modeling uncertainty. 
Those issues could mitigated by 
learning a more accurate representation of the model~\cite{zeng2020tossingbot}, or designing a learned controller explicitly reasoning about dynamic residual between the simulation and the real world~\cite{kaufmann2023champion}.  

Third, our workspace is strictly two-dimensional, as is represented via occupancy grids. While sufficient for planar navigation tasks, it does not capture the full geometry of 3D environments; incorporating richer modalities such as depth images or point clouds~\citep{3Ddiffusion} could address this limitation. 

Fourth, our diffusion model is trained on pre-collected expert demonstrations. This enables straightforward training but limits applicability in environments lacking demonstration data. Extending our approach to learn behaviors from scratch using reinforcement learning~\citep{RLsurvey} is a promising direction.

Lastly, we employ a bare-bone implementation of sampling-based planners that lack optimization and fine-grained parallelism, limiting our ability to fully exploit batch inference in diffusion models. Incorporating recent parallelized frameworks such as ~\citep{prrtc,vamp} could significantly improve runtime efficiency.

\acknowledgments{
We would like to thank Ori Menashe for his valuable assistance with the experiments, and Tomer Michaeli for his insightful comments on  theoretical aspects of diffusion models.
}





\bibliography{main}  

\newpage

\appendix
\section*{Appendix A: Theoretical Guarantees}\label{sec:proof}
\setcounter{section}{1}
In this section, we prove Theorem~\ref{thm:rrt_pc}, namely that DiTree, when using an RRT-based node selection mechanism, is PC, under the assumption that the underlying action sampling mechanism has full support.  
To obtain the proof, we review the main ingredient in the PC proof of the classic RRT algorithm and identify necessary changes to our setting. We then demonstrate that diffusion models satisfy the full support assumption. We conclude the section with a discussion about obtaining AO guarantees when using AO-RRT or SST as underlying SBPs. 

\subsection{Probabilistic Completeness of RRT}
We review the main ingredients in RRT's PC proof~\cite[Theorem~2]{RRTPC}. The proof assumes that (i) the dynamics $f$ are Lipschitz continuous with respect to states and controls~\cite[Definition~2]{RRTPC}, and (ii) that there is a \emph{robust} solution to the KMP problem, i.e., a solution trajectory with positive clearance. (Correspondingly, we make those common assumptions in our Theorem~\ref{thm:rrt_pc} as well.) 

The proof then proceeds in the following manner. Denote by $\pi$ the robust nominal solution trajectory, $T$ its duration, and $\mathbf{u}$ its control function. A set of landmark states $x_0,\ldots, x_m$ is chosen along $\pi$, where $x_i:=\pi(i\cdot \tau)$ and $\tau$ is a constant duration, where $x_0=x_\text{start}$ and $x_m\in \mathcal{X}_\text{goal}$. Then, each landmark is associated with a radius $r_i$-ball centered at $x_i$, denoted by $B_{r_i}(x_i)\subset \mathcal{X}$.  

It is shown that RRT visits each of those balls, with probability at least $1-a'e^{-b'k}$, and thus finds a feasible solution. To achieve this, the proof computes a lower bound on the probability of making progress between two consecutive balls. Specifically, fix $0<i<m$ and suppose that the RRT tree has vertex $x'_i\in B_{r_i}(x_i)$. Then, the probability that RRT obtains a new  vertex $x'_{i+1}\in B_{\kappa r_{i+1}}(x_{i+1})$ in the current iteration, is at least $\rho_i:=p^x_i\cdot p^t_i \cdot p^u_i$, for some $\kappa\in (0,1)$. 

Specifically, the value $p^x_i$ denotes the probability of selecting an RRT node for expansion from $B_{r_i}(x_i)$ (line~4 in Alg.~\ref{alg:SBP}), which is at least the probability of drawing $x_\text{rand}\in B_{r_i2/5}(x_i)$ uniformly at random from $\mathcal{X}$~\cite[Lemma~4]{RRTPC}. Assuming that this event occurs, i.e., $x_\text{near}\in B_{r_i}(x_i)$, the values $p^t_i$ and $p^u_i$ jointly lower-bound the probability of sampling a time duration $t_\text{rand}\in [0,T_\text{prop}]$ and action $u_\text{rand}\in \mathcal{U}$ such that their propagation results in a new vertex $x_\text{new}\in  B_{\kappa r_{i+1}}(x_{i+1})$. In particular, $p^t_i$ denotes the probability that $t_\text{rand}\in [T_-,T_+]$, for some constants $0<T_-\leq T_+ \leq T_\text{prop}$. Similarly, $p^u_i$ denotes the probability that $u_\text{rand}$ is sampled \emph{uniformly at random} from $B_{\Delta u}(u_i)\subset \mathcal{U}$, which is a ball in the control space of radius $\Delta u>0$ centered at $u_i$, which is the nominal control at $x_i$~\cite[Lemma~3]{RRTPC}.  Importantly, the sampled action $u_\text{rand}$ is applied through the entire duration $t_\text{rand}$ of propagation. 

The constants $\tau,m,\kappa,r_i,T_-,T_+$, and $\Delta u$ depend on the nominal trajectory clearance, its duration, and the Lipschitz constants of $f$, but they do not depend on the number of RRT iterations $k$. They are chosen to ensure that the propagation described in the previous paragraph results in a collision free trajectory reaching the next ball. Moreover, those values are carefully selected such that each of the values $p^x_i, p^t_i$ and $p^u_i$ is greater than zero. 

As a result, the probability $\rho_i>0$ holds for any $0\leq i\leq m$. Moreover, as it is independent of $k$, a Markov argument allows to lower-bound the probability that the final ball is eventually reached with $1-a'e^{b'k}$, which concludes the proof of~\cite[Theorem~2]{RRTPC}.

\subsection{Probabilistic Completeness of DiTree}
\label{subsec:ditree_pc}

For our proof of Theorem~\ref{thm:rrt_pc}, it suffices to slightly revise the above reasoning for RRT. In particular, we keep the same construction, but alter the probability of a successful extension from ball $i$ to ball $i+1$, denoted by $\bar\rho_i:=\bar p^x_i\cdot \bar p^t_i \cdot \bar p^u_i$, where the three values correspond in meaning to $p^x_i,  p^t_i$ and $p^u_i$, which we discussed above. 

First, we note that $\bar p^x_i=p^x_i$, which follows from the fact that an RRT-based instantiation of DiTree uses the same node selection mechanism as RRT. 
Second, the choice of $t_{rand}$ remains identical to RRT, therefore $\bar p^t_i=p^t_i$.

Thus, it remains to show that $\bar p^u_i>0$. Recall that rather than  sampling a single control and applying for the entire duration of the propagation, as in RRT, DiTree samples a sequence of $N>0$ controls $u_{\text{rand}_{1:N}} \sim p_\theta\bigl(u_{{1:N}} \mid x_{\text{near}},{x}_{\text{target}}, \mathcal{X}_{\text{obs}}^{\text{near}}\bigr)$ and propagates them sequentially, such that each remains fixed for some $\Delta t>0$, and $N \Delta t=t_\text{rand}$. To mimic the behavior of RRT, we require that for any $1\leq j\leq N$ it holds that $u_{\text{rand}_j}\in  B_{\Delta u}(u_i)$ (adapting Lemma~2 in~\cite{RRTconnect} to this setting is straightforward). That is, 
$\bar p^u_i:=\prod_{j=1}^N \Pr\left(u_j\in B_{\Delta u}(u_i)\right)$, which is greater than zero if $p_\theta\bigl(u_{{1:N}} \mid x_{\text{near}},{x}_{\text{target}}, \mathcal{X}_{\text{obs}}^{\text{near}}\bigr)$  
has full support. 
This concludes the proof of Theorem~\ref{thm:rrt_pc}.

\subsection{Full Support of Diffusion and Flow Matching}
\label{subsec:diff_full_support}
Next, we justify our full support assumption, i.e., given that 
$u_{\text{rand}_{1:N}}\sim p_\theta\bigl(u_{\text{rand}_{1:N}} \mid x_{\text{near}},{x}_{\text{target}}, \mathcal{X}_{\text{obs}}^{\text{near}}\bigr)$
, where the distribution is according to a diffusion model (DMs)~\citep{diffusion,DDIM} or flow matching models (FMs)~\citep{lipmanflowm} with weights $\theta$, then $u_{\text{rand}_{j}}\in B_r(\cdot )$ for all $1\leq j\leq N$, where the ball is of radius $r>0$ centered around some action in $\mathcal{U}$. 
For simplicity, we denote the distribution from which we sample as $p_\theta(u)$, where $u:=u_{\text{rand}_{1:N}}$. We also assume that the distribution is unconditioned (the derivation for conditional distributions such as DiTree's $p_\theta\bigl(u_{\text{rand}_{1:N}} \mid x_{\text{near}},{x}_{\text{target}}, \mathcal{X}_{\text{obs}}^{\text{near}}\bigr)$ is straightforward and follows directly from prior work~\citep{diffusionSDE,DDIM,lipmanflowm}, without affecting the core argument). Without loss of generality, we consider the setting of $N=1$, as the arguments below can be easily generalized by considering a diffusion model whose domain is $\mathbb{R}^{ND}$, rather than $\mathbb{R}^D$, which we consider below. 


We provide necessary mathematical details on DMs and FMs and their sampling process, and then discuss their full support property. Specifically, we focus on the inner workings of generative sampling by integration\footnote{We emphasize this integration is \emph{not} related to the integration of the dynamic model $f$ used in DiTree's forward propagation (Line~5 Alg.~\ref{alg:SBP}).} of an SDE or ODE. Both DM and FM aim to learn a transformation
$
T_\theta: \mathbb{R}^D \to \mathbb{R}^D
$
that maps a sample $u_0$ from a distribution $\mathcal{N}(0, I)$\footnote{While FM could use other source distributions, in practice the normal distribution is commonly chosen.} to a sample $u_1 \sim p_{\text{data}}$ in a complex target distribution, which in DiTree is the conditional distribution of expert action sequences. This transformation is defined over a time interval $t \in [0, 1]$ by integrating a continuous, time-dependent vector field $\mathbf{v}_\theta(u_t, t)$, parameterized by $\theta$, that governs the evolution of the state $u_t \in \mathbb{R}^D$. The boundary conditions for this evolution are
$$
u(0) = u_0, \quad u(1) = u_1 = T_\theta(u_0).
$$

The nature of the vector field $\mathbf{v}_\theta(u_t, t)$ differs between the two paradigms. In DMs, $\mathbf{v}_\theta(u_t, t)$ entails an approximation of the score function, i.e., the gradient of the log-density of the time-dependent marginal $p_t(u)$:
  $$
  \mathbf{v}_\theta(u_t, t) \propto \nabla_u \log p_t(u_t).
  $$


In FMs, $\mathbf{v}_\theta(u_t, t)$ is a velocity field that directly defines the transport direction and magnitude needed to continuously morph the source distribution into the target distribution across time.

In both cases, the overall transformation $T_\theta(u_0) = u_1$ is realized by solving the differential equation
$$
\frac{ \dd u(t)}{ \dd t} = \mathbf{v}_\theta(u_t, t),
$$
or its stochastic counterpart, depending on the formulation.
The choice between a deterministic ordinary differential equation (ODE)~\citep{DDIM} or a stochastic differential equation (SDE)~\citep{diffusionSDE} determines whether the generative process is deterministic or includes sampling noise.


We now detail the full support property of sampling under both the ODE and SDE formulations.
Our explanation relies on the sampling dynamics themselves using a numerical solver and a learned vector field, rather than whether the specific training method used is DM or FM.
Therefore, the justification below applies to both types of models, and we use them interchangeably.


\subsubsection{ODE Formulation}
In the deterministic case, the generative process is described by the associated probability flow ODE (e.g., as in~\citep{DDIM}).
Sampling $p_\theta(u)$ is  performed by integrating from an initial Gaussian sample using a numerical solver. The time interval $[0,1]$ is discretized as \( t_0 < t_1 < \cdots < t_\ell \) with a time step $\Delta t$ and the Euler method is applied as a numerical solver for integration over  $\mathbf{v}_\theta(u_t,t)$. 
The value $u_{t_{i+1}}$ is derived from the previous value by the update
\begin{align}
    u_{t_{i+1}} = u_{t_{i}} + \mathbf{v}_\theta(u_{t_{i}},t_i)\Delta t. \label{eq:ode}
\end{align}

Next, we show that $u_1=u_{t_{\ell}}$ has full support on $\mathbb{R}^D$. 
Let \( u_0 \sim \mathcal{N}(0, I) \) denote the initial sample drawn from the source Gaussian, and let \( T_\theta: \mathbb{R}^D \to \mathbb{R}^D \) be the transformation defined by integrating a 
vector field \( \mathbf{v}_\theta(u_t, t) \) over time \( t \in [0, 1] \), and \( u_1 = T_\theta(u_0) \) the generated output. Since the vector field $\mathbf v_\theta(u,t)$ is globally Lipschitz and continuously differentiable, it yields a unique $C^{1}$ flow map $T_\theta$  via the Picard--Lindelöf theorem~\citep{arnoldODEs}, which is a diffeomorphism onto its image.

To conclude, since the Gaussian distribution assigns positive probability to every non-empty open subset of \( \mathbb{R}^D \), for any non-empty open set \( B_r(\cdot ) \subset \mathbb{R}^D \), the preimage \( T_\theta^{-1}(B_r(\cdot )) \) is also open and has positive measure under \( u_0 \).



\subsubsection{SDE Formulation}

The SDE formulation introduces sampling noise to the ODE (see, e.g.,~\citep{diffusionSDE}), where 
the Euler--Maruyama method is applied to numerically simulate the SDE, which yields the update rule 
%
\[
{u}_{t_{i+1}} = \underbrace{{u}_{t_i} + v_\theta({u}_{t_i}, t_i)\Delta t}_{\text{deterministic shift}} + \underbrace{g(t_i)\sqrt{|\Delta t|} \, \epsilon_i}_{\text{Gaussian noise}},
\]
where \( g(t) \) is the diffusion coefficient---a scalar-valued function controlling the noise magnitude at time $t$.
It must satisfy $g(t)>0$ for $t>0$ to ensure well-defined noise injection.
The noise term \( \epsilon_i \sim \mathcal{N}(0, I) \) is a Gaussian random variable with zero mean and full-rank (positive-definite) covariance matrix, simulating the increment of a standard Wiener process (Brownian motion) over the time interval \( \Delta t \). 

The final update step in the sampling process ${u}_{\ell} = {u}_{\ell-1} + v_\theta({u}_{\ell-1}, t_{\ell-1})\Delta t + g(t_{\ell-1})\sqrt{|\Delta t|} \, \epsilon_{\ell-1}$ is a summation of a deterministic term and a Gaussian noise term, which results in a Gaussian random variable with a shifted mean and non-degenerate covariance. Similarly to the ODE setting, this implies the full support of the distribution.

\textbf{Remark.}
In other generative methods where the full support assumption might
not strictly hold---for instance, if the dynamics become partially
deterministic, or if the noise coefficient $g(t)$ vanishes at some
positive times---full support can still be recovered by adding a small
amount of independent noise to the final sample.
Specifically, perturbing the output by a small Gaussian
variable ensures that the resulting distribution has a strictly positive
density everywhere.

\subsection{Extension to Asymptotic Optimality}
\label{subsec:aorrt}
Our theoretical guarantees of DiTree can be further strengthened by substituting the RRT backbone planner with a planner whose solution quality is guaranteed to converge to the optimum---a property called asymptotic optimality (AO).

While RRT is PC under the assumptions of Lipschitz dynamics and the existence of a robust trajectory, it does not offer any guarantees about the \emph{quality} of the returned path. In particular, RRT may return highly suboptimal solutions even with an infinite number of samples~\cite{NechushtanRH10}. 
To address this limitation, the AO-RRT~\cite{statecostspaceRRT} was proposed.
The central idea in AO-RRT is to augment the original state space $\mathcal{X}$ with an additional cost dimension. That is, AO-RRT operates in the $(d+1)$-dimensional space $\mathcal{Y} := \mathcal{X} \times \mathbb{R}_{\ge 0}$, where each point $y = (x, c)$ stores both the state $x$ and the cost-to-come $c$ from $x_\text{init}$ to $x$. The system dynamics are lifted to this space via the augmented dynamics
\[
\dot{y} = F(y,u) = (f(x,u),\,g(x,u)),
\]
where $g(x,u)$ is the instantaneous cost function and is assumed to be Lipschitz continuous, like $f$. With this formulation, AO-RRT can be viewed as RRT operating over the dynamics $F$ in the space $\mathcal{Y}$. Thus, the AO proof of AO-RRT~\cite{AOAnalysis} follows the same reasoning as the PC proof of RRT~\cite{RRTPC}. Thus, by relying on the backbone of AO-RRT in DiTree, the AO property will also extend to DiTree, due to the full support of DM and FM. 

Another planner that could be used in conjunction with DiTree to achieve AO is SST~\citep{SST}. Although the structure of the AO proof of SST differs from that for AO-RRT, both algorithms rely on very similar building blocks. However, some care should be taken when revisiting all the steps of SST's proof, which we leave for future work.

\section*{Appendix B: Robot Dynamics}
\label{sec:car_dynamics}
We provide additional details on the robot dynamics. 
The car dynamics are based on a single-track dynamic model used in~\citep{carModel}. This well-established model offers a good balance between physical realism and computational tractability. Compared to simplified models such as unicycle, Ackermann steering, or even kinematic bicycle models, the single-track model captures key inertial and actuation constraints. Specifically, it captures essential second-order effects such as acceleration saturation and steering inertia, which are critical for planning under dynamic constraints. 
 
The state space is defined as $\{x, y, \psi, v, D, \delta\}$, where $\{x, y\}$ denote position, $\psi$ the heading angle and $\delta$ the front wheel steering angle. The action space is $\{\dot{D}, \dot{\delta}\}$, representing throttle and steering rate commands.
 System dynamics are given by:
\begin{align*}
    \dot{x} &= v \cos(\psi + c_1 \delta), \\
    \dot{y} &= v \sin(\psi + c_1 \delta), \\
    \dot{\psi} &= v c_2 \delta, \\
    \dot{v} &= \frac{F_x}{m} \cos(c_1 \delta),
\end{align*}
where the longitudinal force $F_x$ is modeled as:
\begin{equation*}
    F_x = (c_{m1} - c_{m2} v) D - c_{r2} v^2 - c_{r0} \tanh(c_{r3} v).
\end{equation*}
Here, $m$ denotes the vehicle mass, $c_{m1}$ is the motor gain, and $c_{m2}$ captures velocity-dependent damping. The terms $c_{r0}$ and $c_{r2}$ model static and quadratic rolling resistance, respectively. 

\begin{figure}[b]
    \centering
    \includegraphics[width=0.4\linewidth]{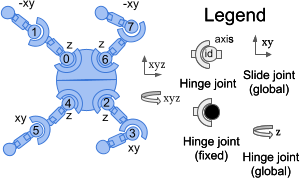}
    \caption{A schematic of the ant robot, borrowed from~\citep{gymnasium}.}
    \label{fig:ant}
\end{figure}

Mujoco's~\citep{mujoco} ant robot is a quadruped robot consisting of four legs, each composed of two segments connected via hinge joints (Fig.~\ref{fig:ant}). The upper segment of each leg connects to the torso with an additional hinge joint. The state space is 29-dimensional, comprising all joint angles, as well as the position and velocity of the ant's torso. The action space is 8-dimensional, representing the torques applied to each joint.





\section*{Appendix C: Additional Implementation Details}


This section provides additional information regarding DiTree's implementation and experiments.

\textbf{Data Collection.} We train each robot on a dataset of expert trajectories. While the ant robot has an available dataset in D4RL~\citep{gym_robotics,Minari}, CarMaze is not part of the D4RL suite. Thus, we generate a dataset by first planning paths using A* on a discretized grid and then executing them using a PD controller to compute dynamically feasible trajectories. Collision-free trajectories were gathered into a dataset.

\textbf{Validation.} Evaluating the diffusion model's generalization during training was performed by executing rollouts on a separate validation set (distinct from the test set) and analyzing performance based on the number of collisions, distance to the goal after $T$ timesteps, and an overall qualitative assessment of path quality and diversity.

\textbf{Nearest Neighbor (NN).} As part of the node selection step (Line~3 of Alg.~\ref{alg:SBP}), RRT uses NN queries. In DiTree's RRT backbone implementation, NN queries are performed using SciPy’s~\citep{SciPy} KDTree.
For our AntMaze tasks only, we chose our distance metric to exclusively account for $(x,y)$ distance to encourage exploration and not be overshadowed by the other $27$ dimensions.

\textbf{Collision Checking (CC)} is not explicitly available in D4RL environments. To enable CC for all the compared methods in our experiments we implement a simplified CC routine by modeling robots as sets of spheres and checking for intersections with neighboring occupancy grid cells. For experiments with OMPL, we use the available Python bindings and set state validation to use the above CC routine. We also configure OMPL to use the environments' \texttt{step()} function for forward propagation, as used by all other methods. 

\section*{Appendix D: Full Experimental Results}
\label{sec:full_results}
\setcounter{section}{4}
\setcounter{subsection}{0}

\subsection{Real-World Experiment}

To evaluate our method in a real-world setting, we constructed a test environment simulating a common racing scenario: turning a corner on a track. A corresponding map was designed to reflect this geometry.

We generated 10 trajectories using both DiTree and a baseline RRT planner, omitting DP and SST since their behaviors are qualitatively similar to DiTree and RRT, respectively. During planning, we used the dynamic car model described in~\citep{carModel} without performing system identification. That is, we adopted the default parameters provided by the implementation. While this choice makes the resulting trajectories harder to track due to model mismatch, it more accurately reflects real-world deployment conditions, where accurate system identification is often infeasible—especially as vehicle dynamics and track conditions vary over time. All other planning and environment parameters were kept unchanged compared to the simulated experiments.

The resulting trajectories were smoothed using a cubic spline interpolation to produce smooth reference paths suitable for real-world execution. A standard pure-pursuit controller was used for trajectory tracking on the physical platform.
Real-time state estimation was provided by an OptiTrack motion capture system. 

Our experiments show classical SBPs trajectories are more difficult to track, which often resulted in collisions. During execution of RRT trajectories, 8 trials out of 10 experienced at least one collision with an obstacle. Collisions were either due to controller overshoot near tight maneuvers or tracking errors when driving in close proximity to obstacles. DiTree trajectories on the other hand had no collisions. 

In addition to number of collisions, we also measure tracking deviation by calculating the distance from each point during execution to the nearest point on the planned path. We find RRT shows a larger average deviation of $0.5[m]$, compared with only $0.28[m]$ for DiTree. A full recording of the experiment can be found in the supplementary video.

\clearpage
\subsubsection{Full Results for Real-World Experiment}
\begin{figure}[!htbp]
    \centering
    \includegraphics[width=1\linewidth]{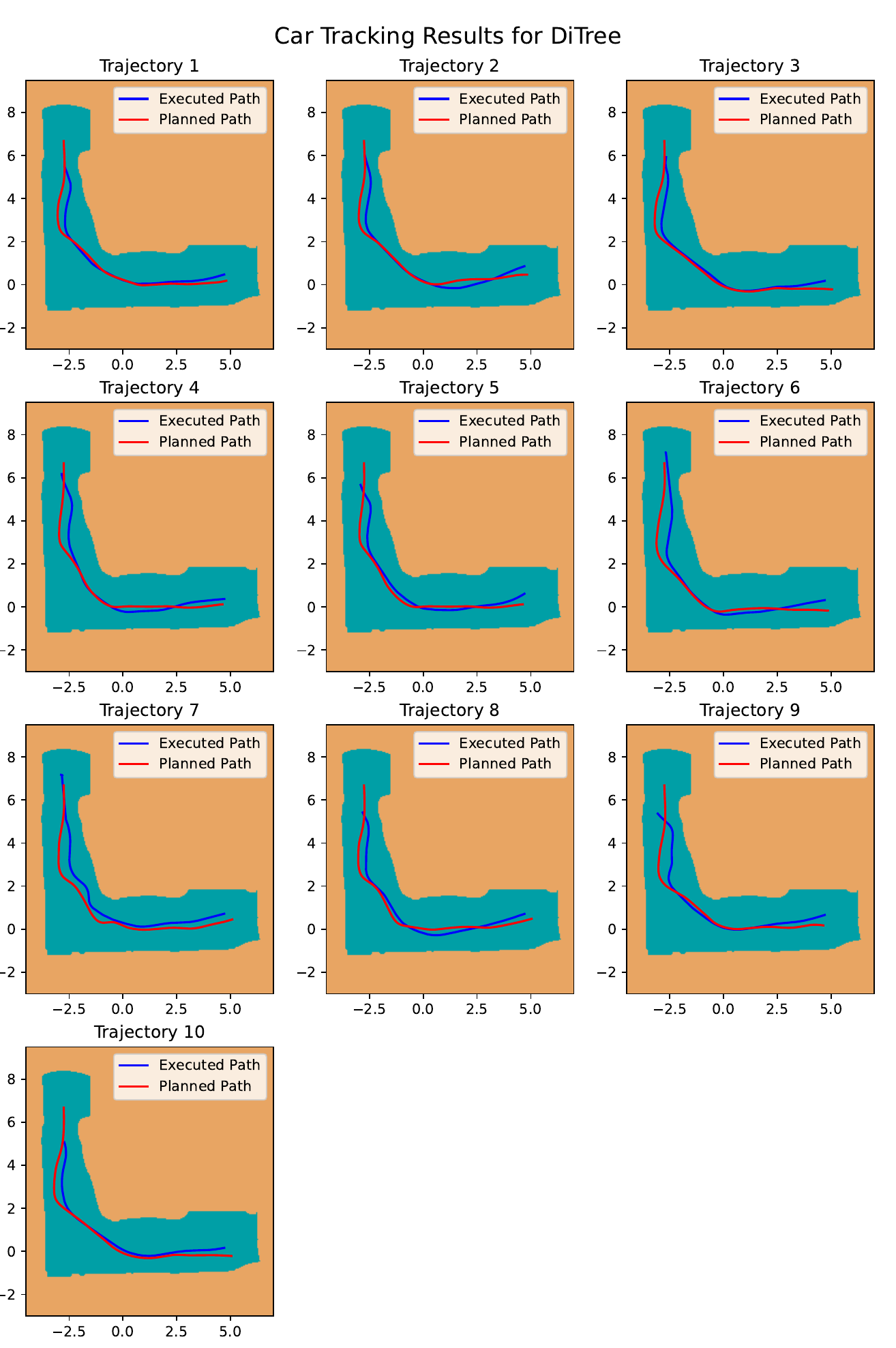}
    \caption{Real car tracking results for DiTree. All axes are expressed in meters.}
    \label{fig:realDiTree}
\end{figure}

\begin{figure}[!htbp]
    \centering
    \includegraphics[width=1\linewidth]{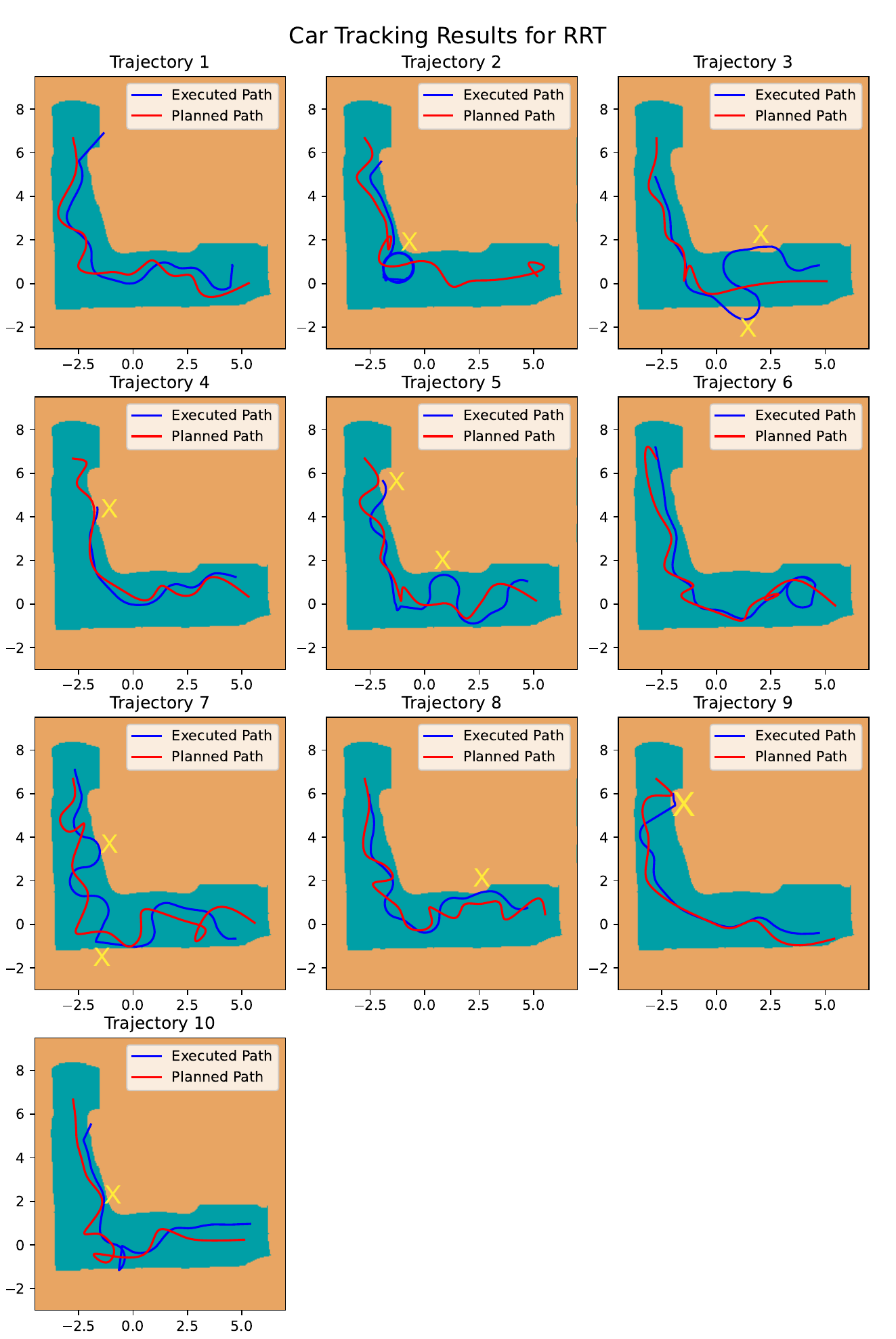}
    \caption{Real car tracking results for RRT, where collisions are marked by a yellow 'X'. Only Trajectories 1 and 6 avoid collisions during execution. Impact with an obstacle while following a loop during trajectory 2 caused a tracking issue. All axes are expressed in meters.}
    \label{fig:realRRT}
\end{figure}

\clearpage
\subsubsection{Full Results for Main Experiment}
\label{subsection:full_results}



\textbf{Computational Bottleneck.} Traditional planners such as RRT and SST rely heavily on low-level geometric operations, issuing approximately 798.6k collision checks per minute of planning in a typical scenario. In contrast, DiTree reduces this number to just 10.5k collision checks per minute. Instead of spending computation time on frequent collision evaluations and nearest-neighbor queries, DiTree allocates the majority of its runtime—94.34\%—to diffusion model inference. This represents a clear reallocation of computational resources from classical search primitives to learned, amortized inference, reflecting a distinct change in the planner’s computational profile. While our bare-bones implementation already exhibits good performance, these results also suggest significant potential for further efficiency gains through model distillation, caching, and other inference-time optimizations~\citep{PruneAndQuant} that could substantially reduce the cost of diffusion-based planning.

Below, we provide full details on the experimental results reported in the main text. 

\begin{figure}[H]
    \centering
    \includegraphics[width=\textwidth]{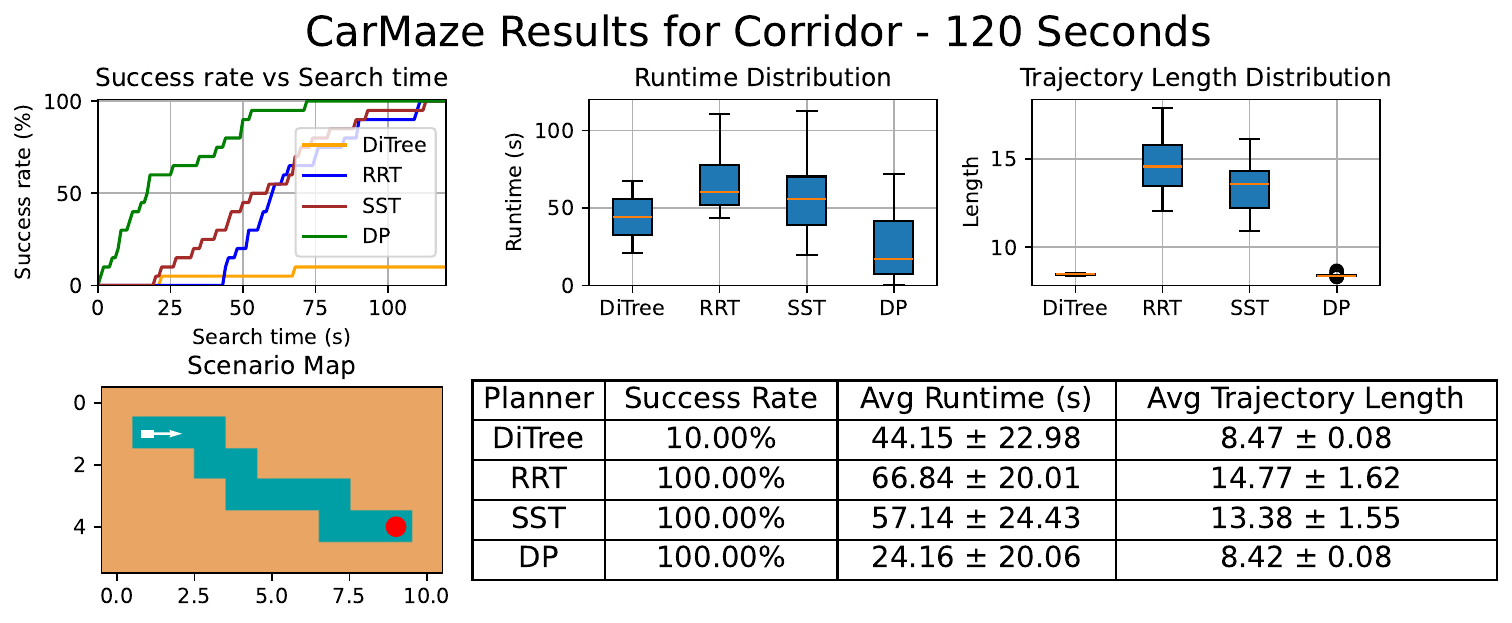}
\end{figure}

\begin{figure}[H]
    \centering
    \includegraphics[width=\textwidth]{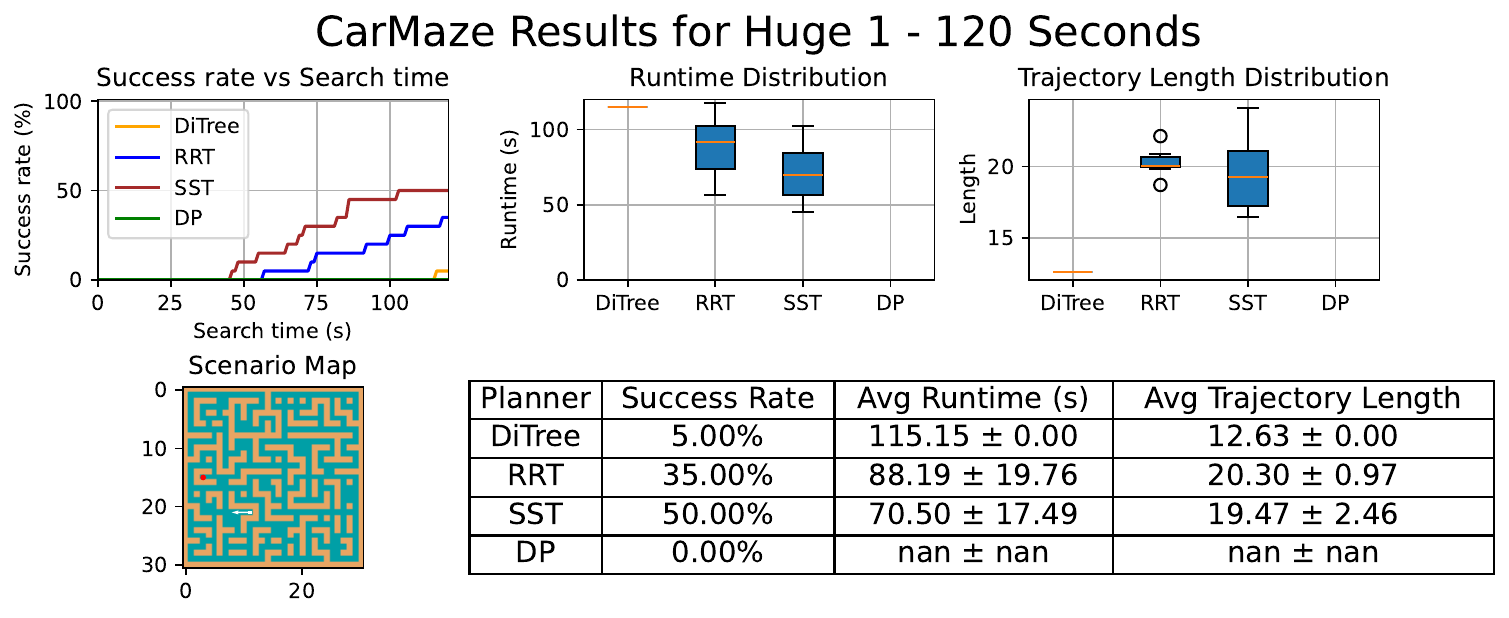}
\end{figure}

\begin{figure}[H]
    \centering
    \includegraphics[width=\textwidth]{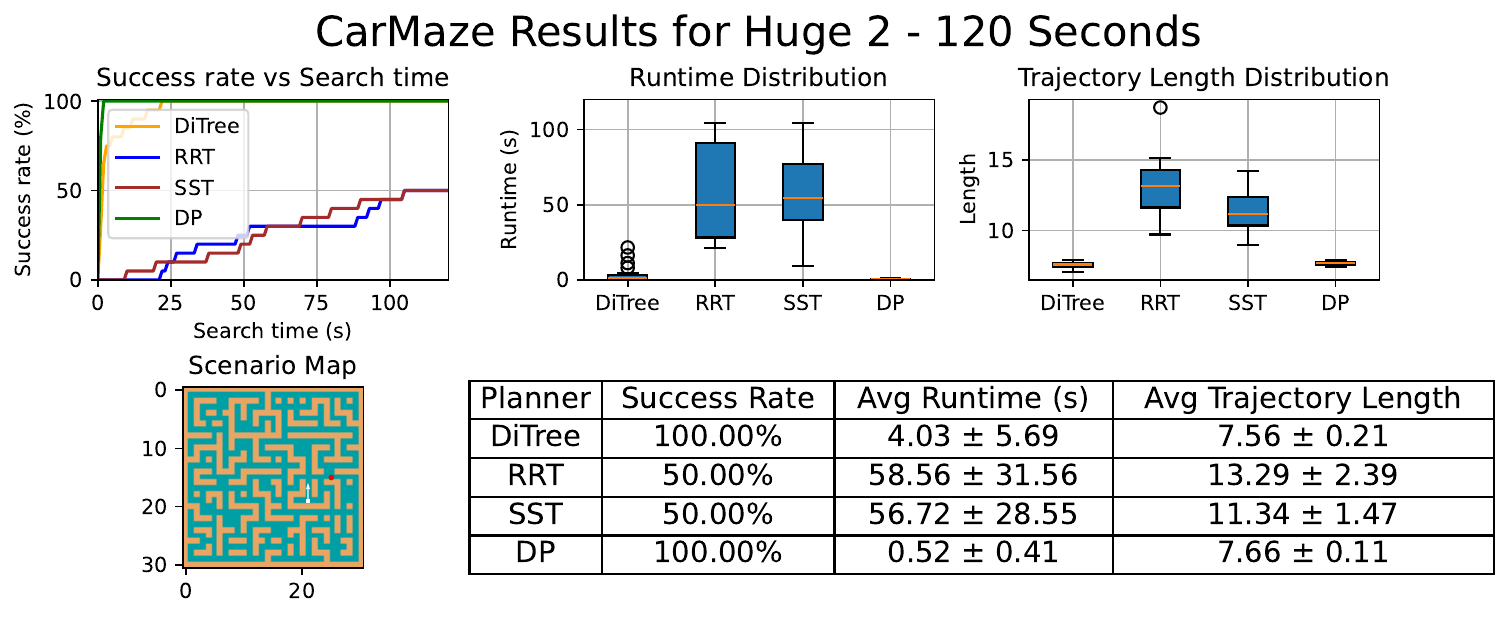}
\end{figure}

\begin{figure}[H]
    \centering
    \includegraphics[width=\textwidth]{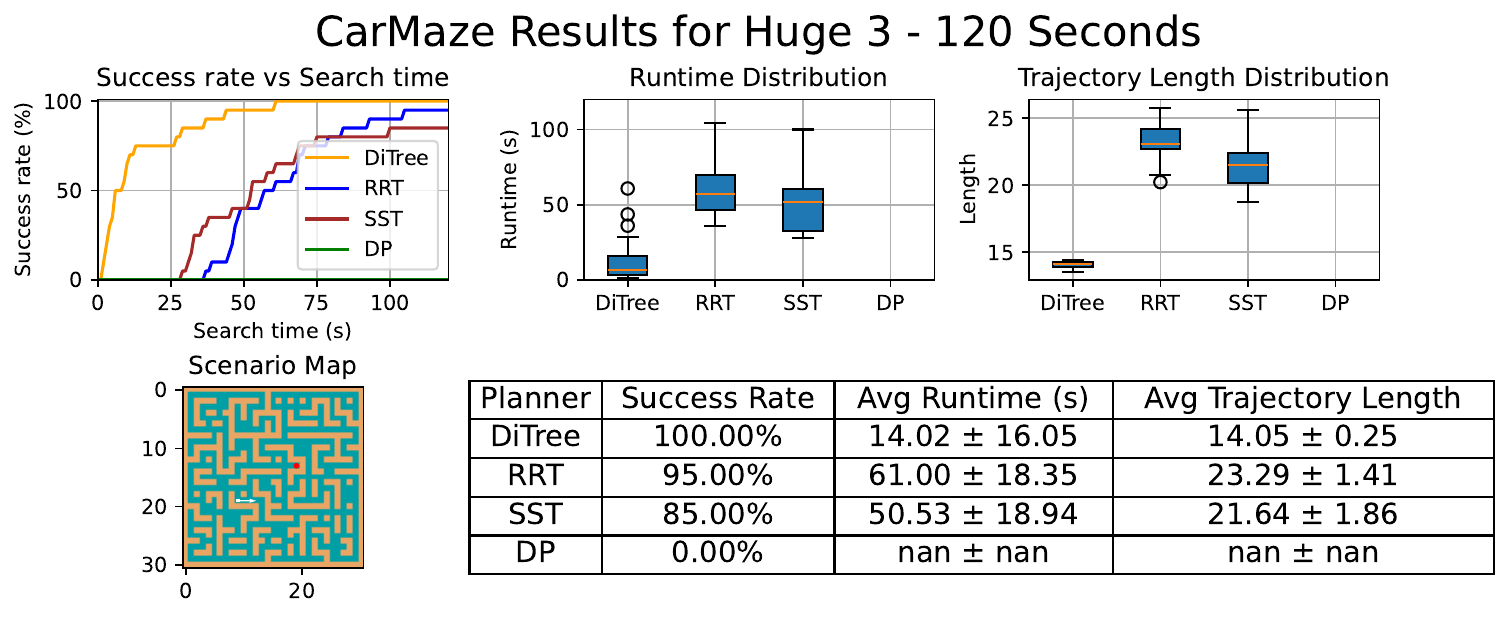}
\end{figure}

\begin{figure}[H]
    \centering
    \includegraphics[width=\textwidth]{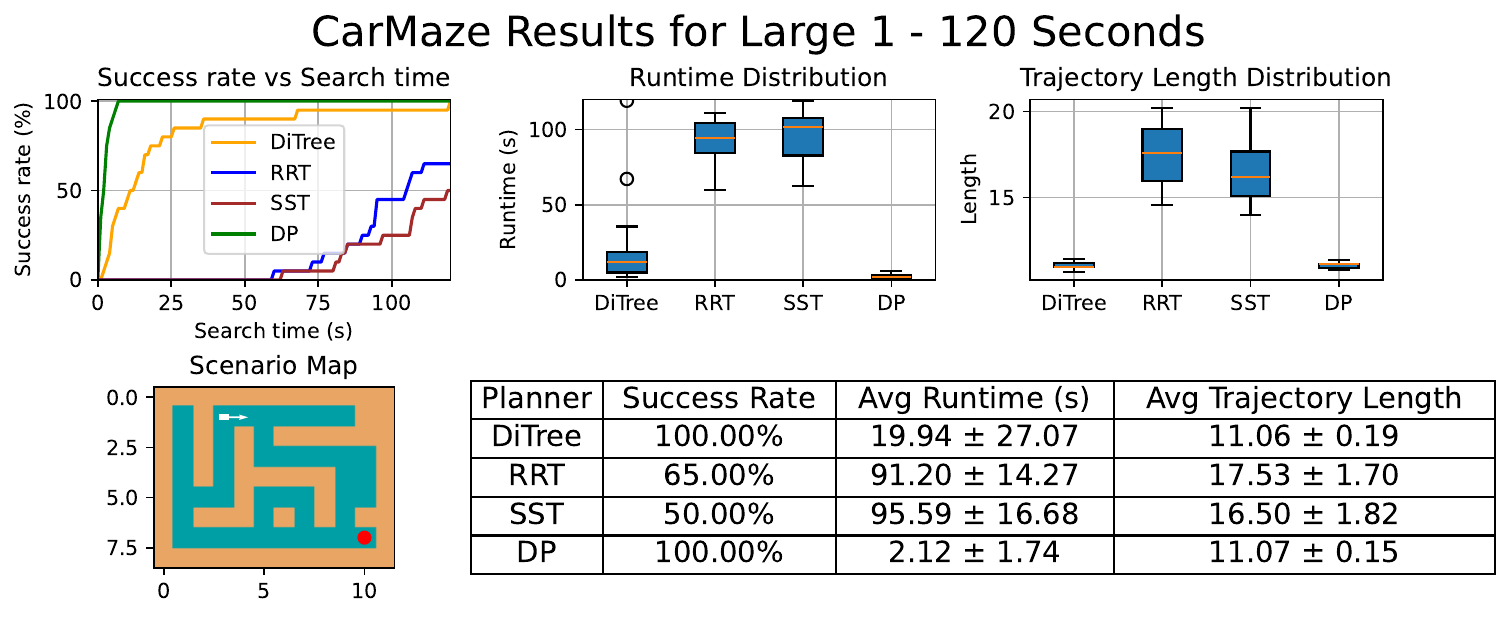}
\end{figure}

\begin{figure}[H]
    \centering
    \includegraphics[width=\textwidth]{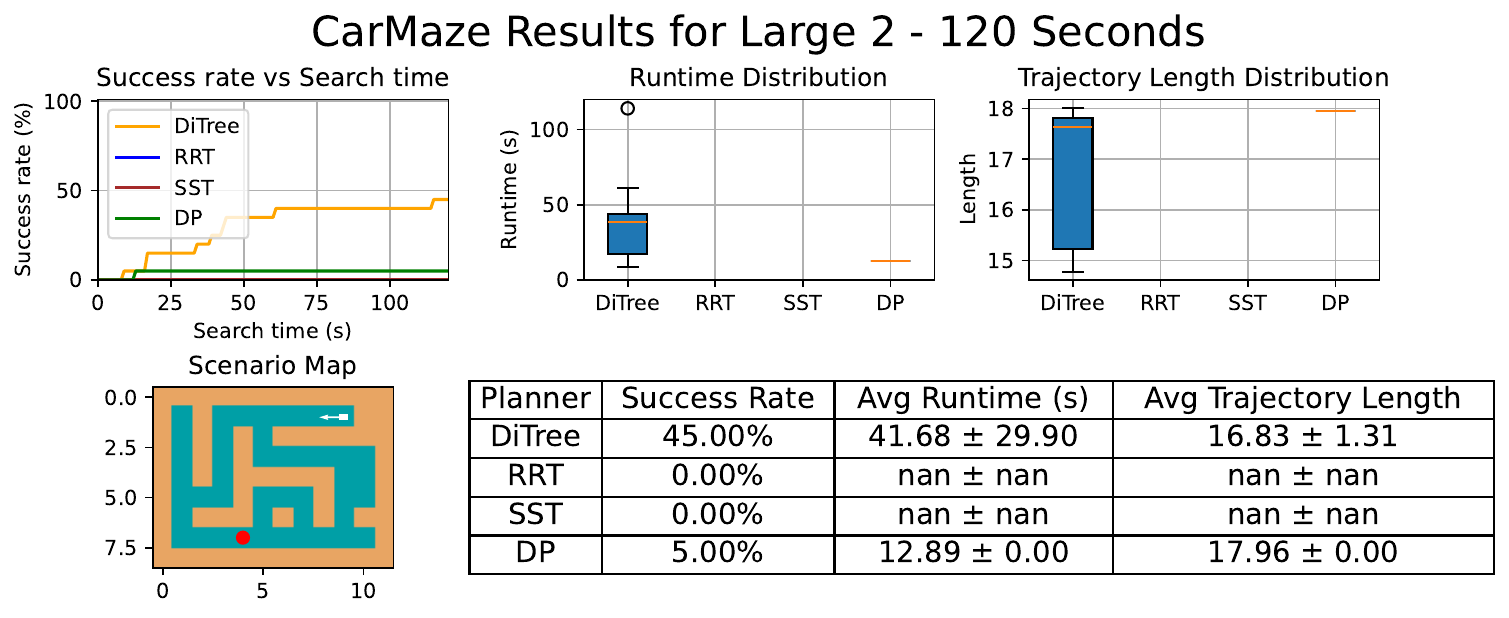}
\end{figure}

\begin{figure}[H]
    \centering
    \includegraphics[width=\textwidth]{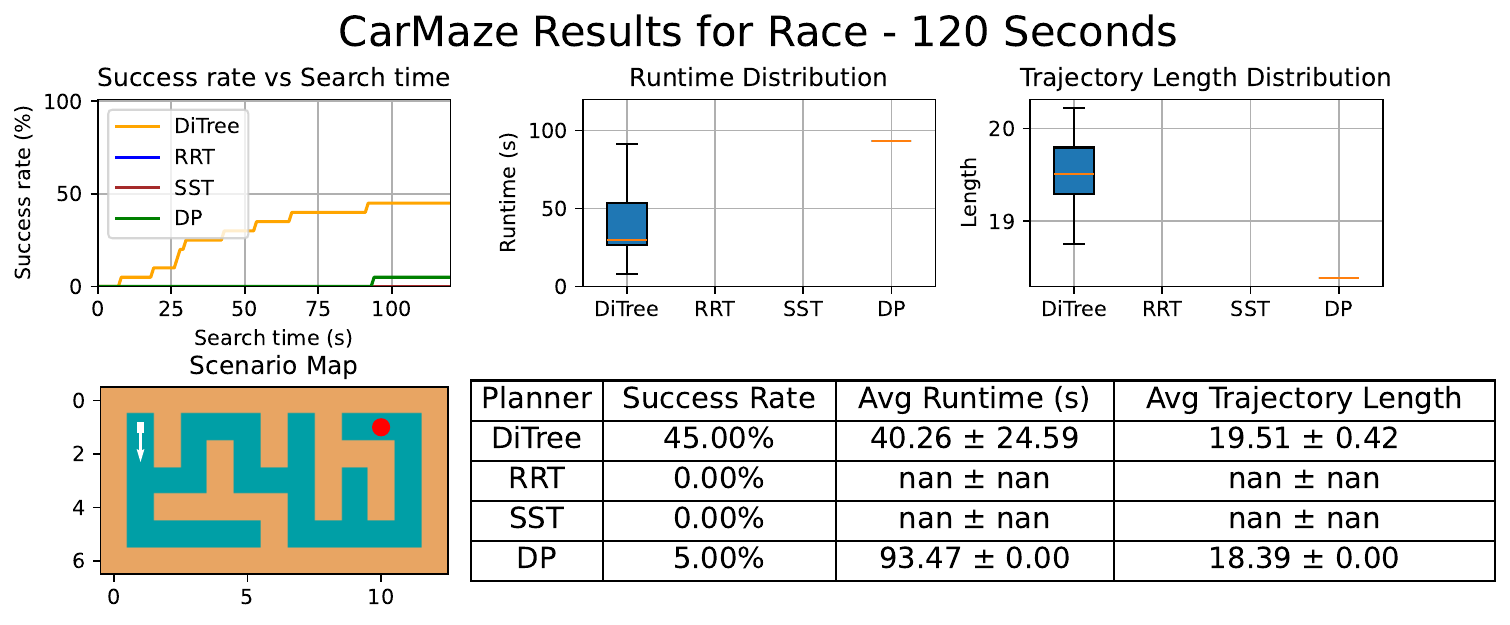}
\end{figure}

\begin{figure}[H]
    \centering
    \includegraphics[width=\textwidth]{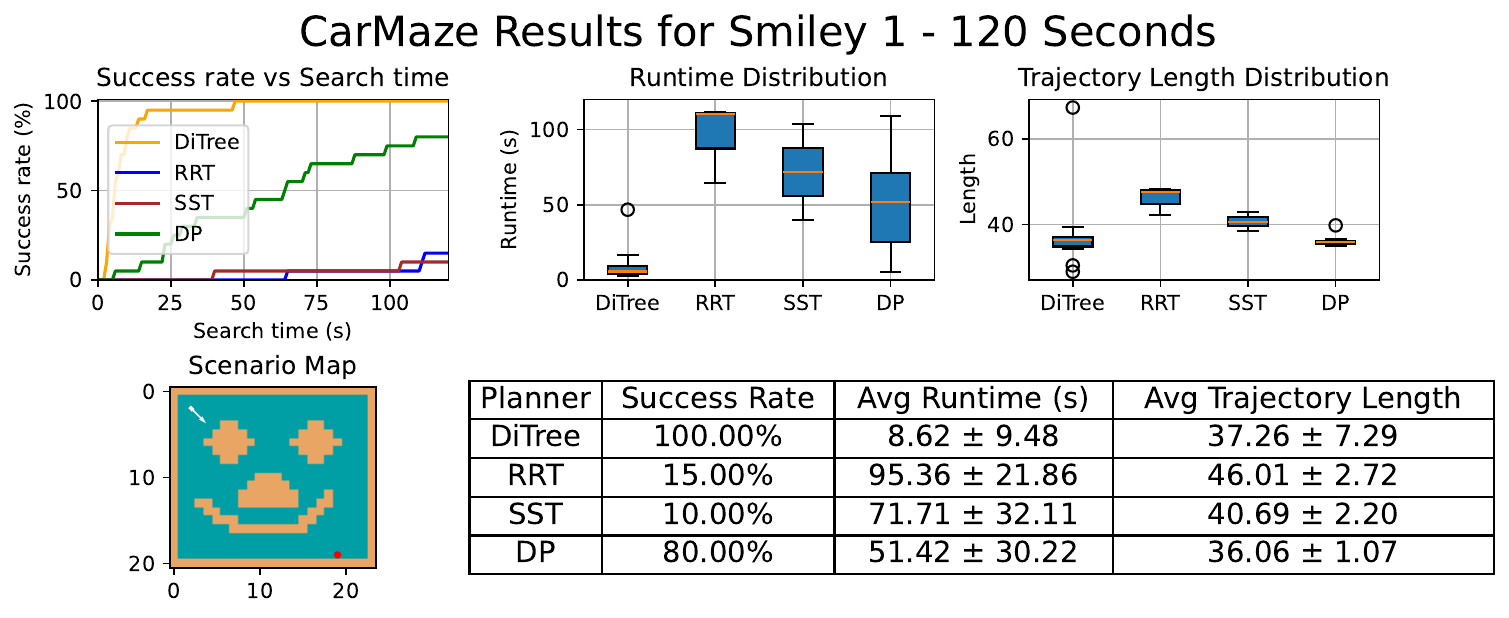}
\end{figure}

\begin{figure}[H]
    \centering
    \includegraphics[width=\textwidth]{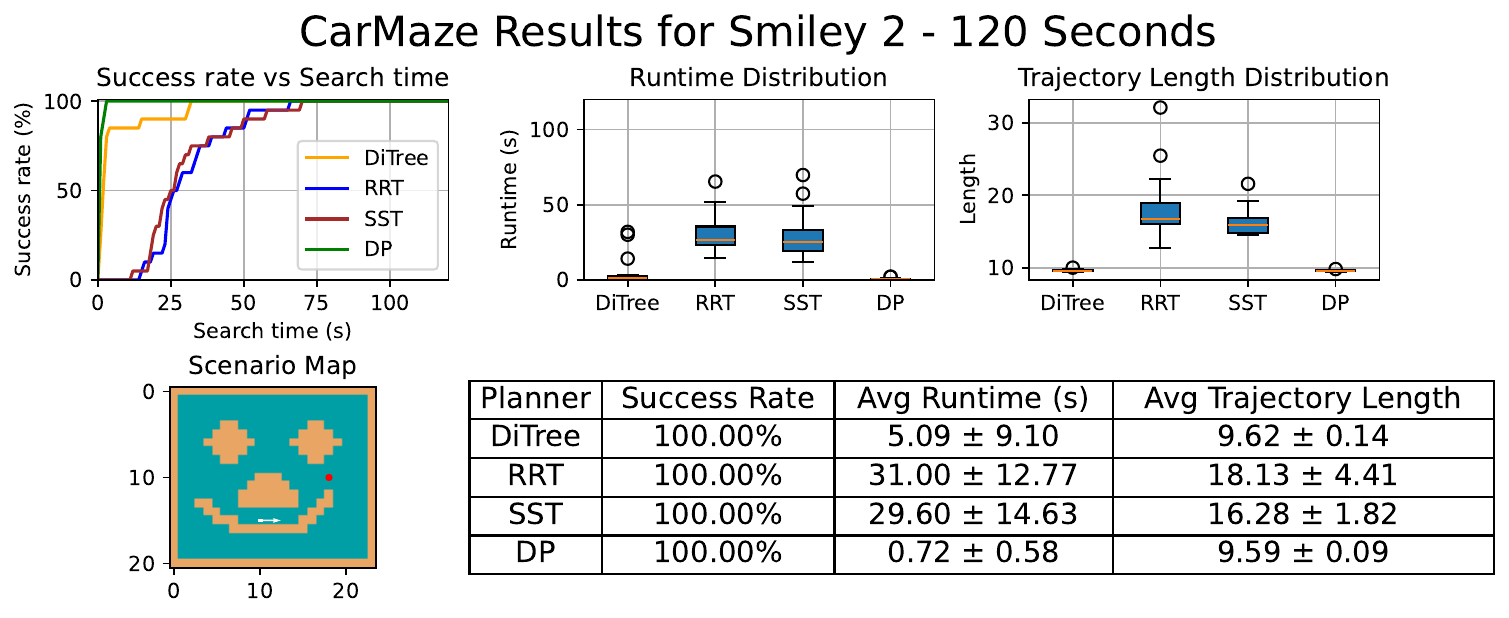}
\end{figure}

\begin{figure}[H]
    \centering
    \includegraphics[width=\textwidth]{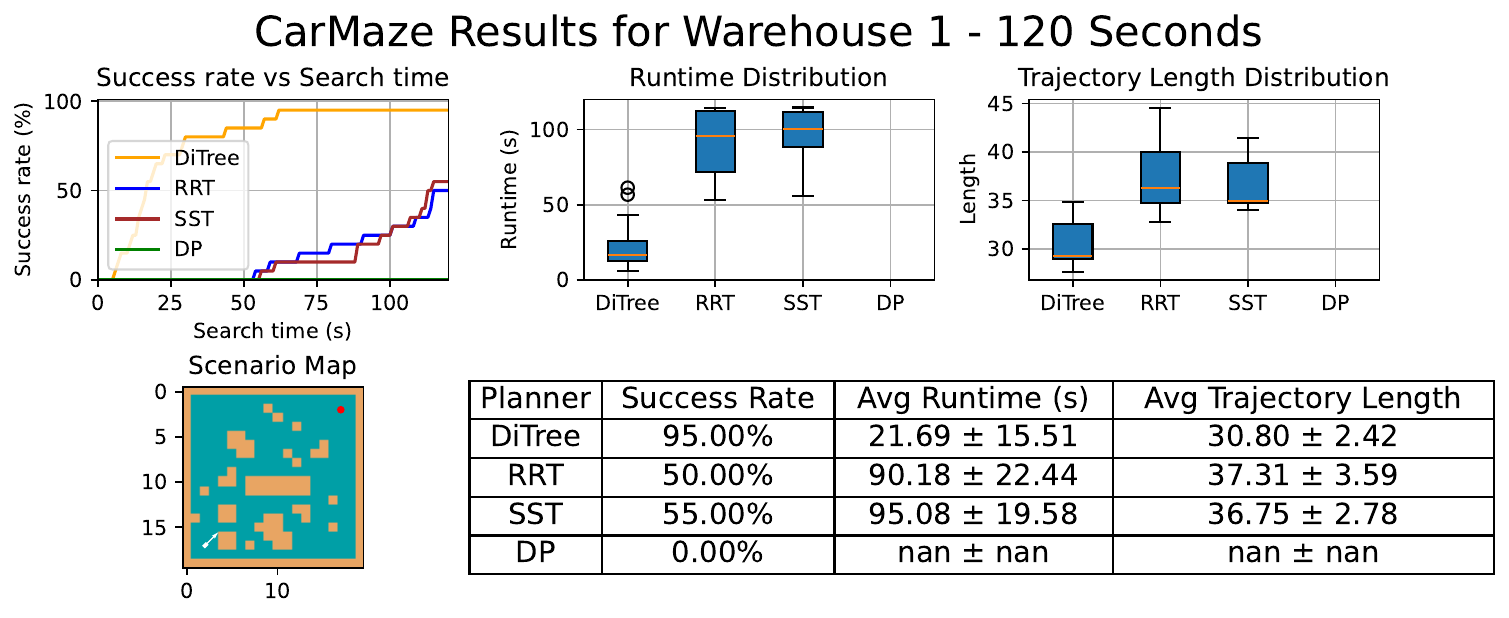}
\end{figure}

\begin{figure}[H]
    \centering
    \includegraphics[width=\textwidth]{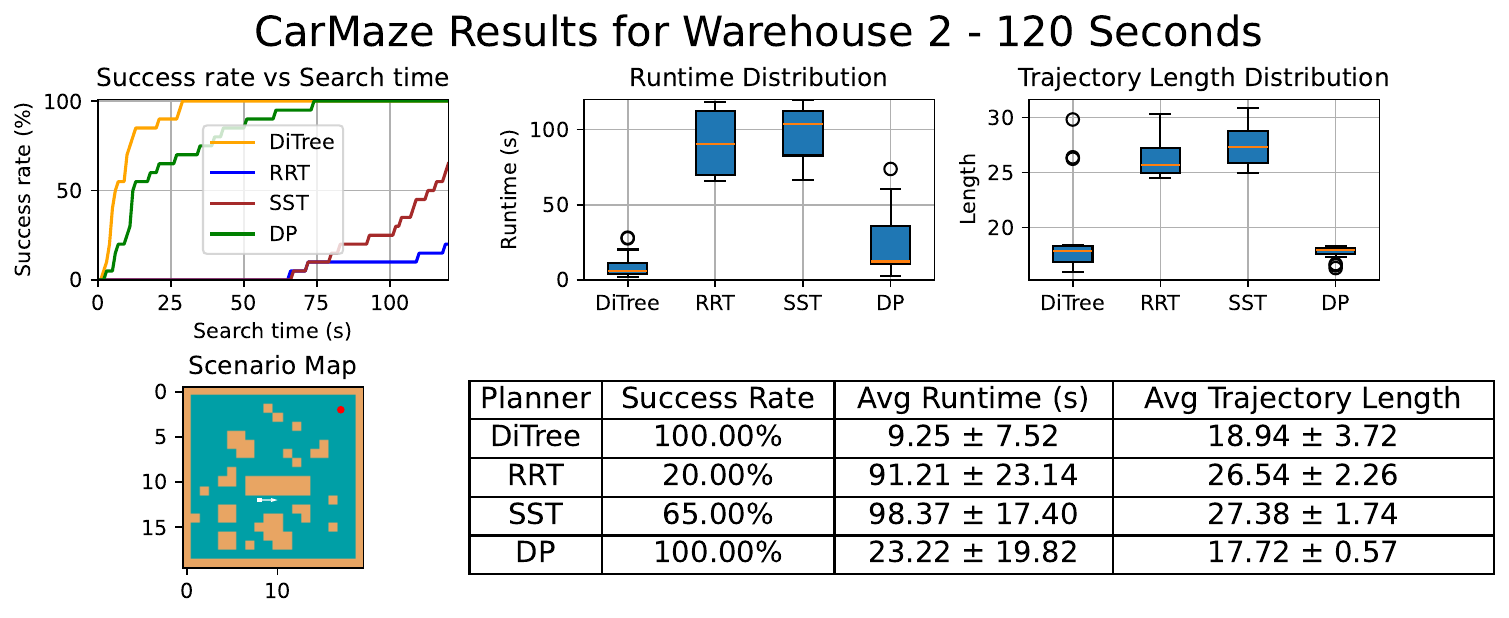}
\end{figure}

\begin{figure}[H]
    \centering
    \includegraphics[width=\textwidth]{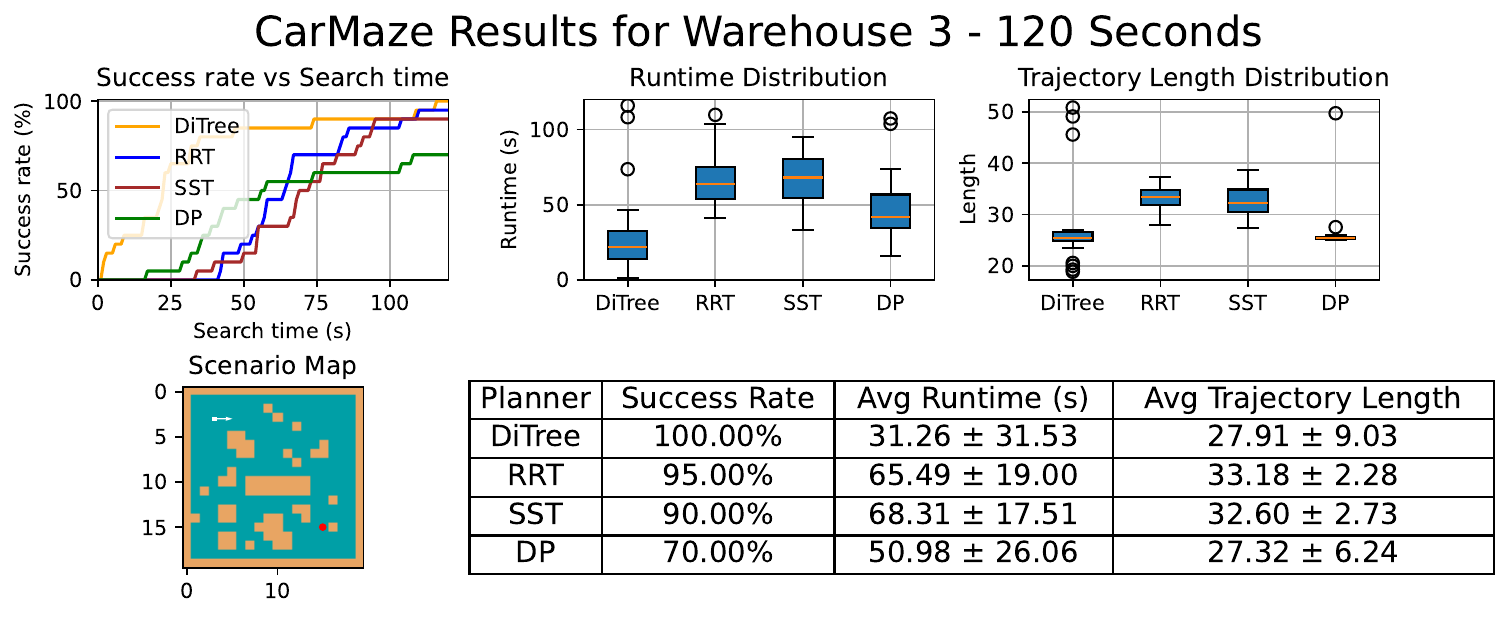}
\end{figure}

\begin{figure}[H]
    \centering
    \includegraphics[width=\textwidth]{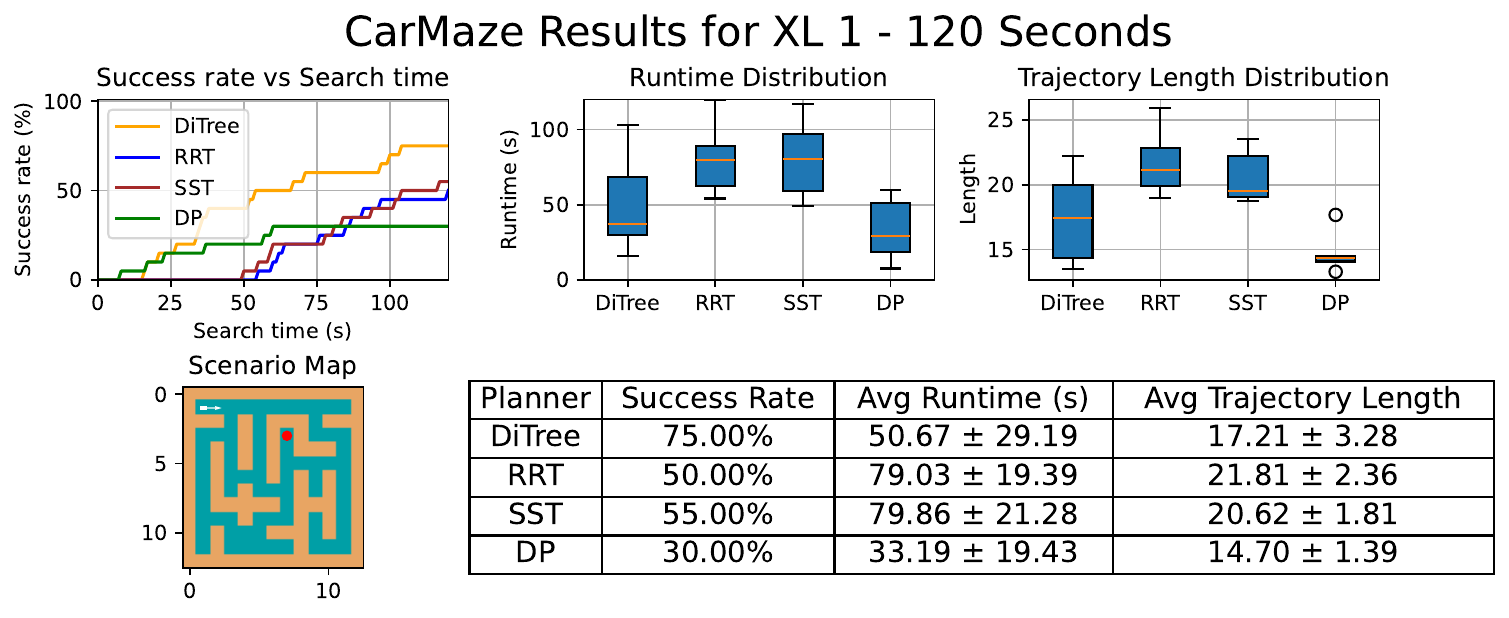}
\end{figure}

\begin{figure}[H]
    \centering
    \includegraphics[width=\textwidth]{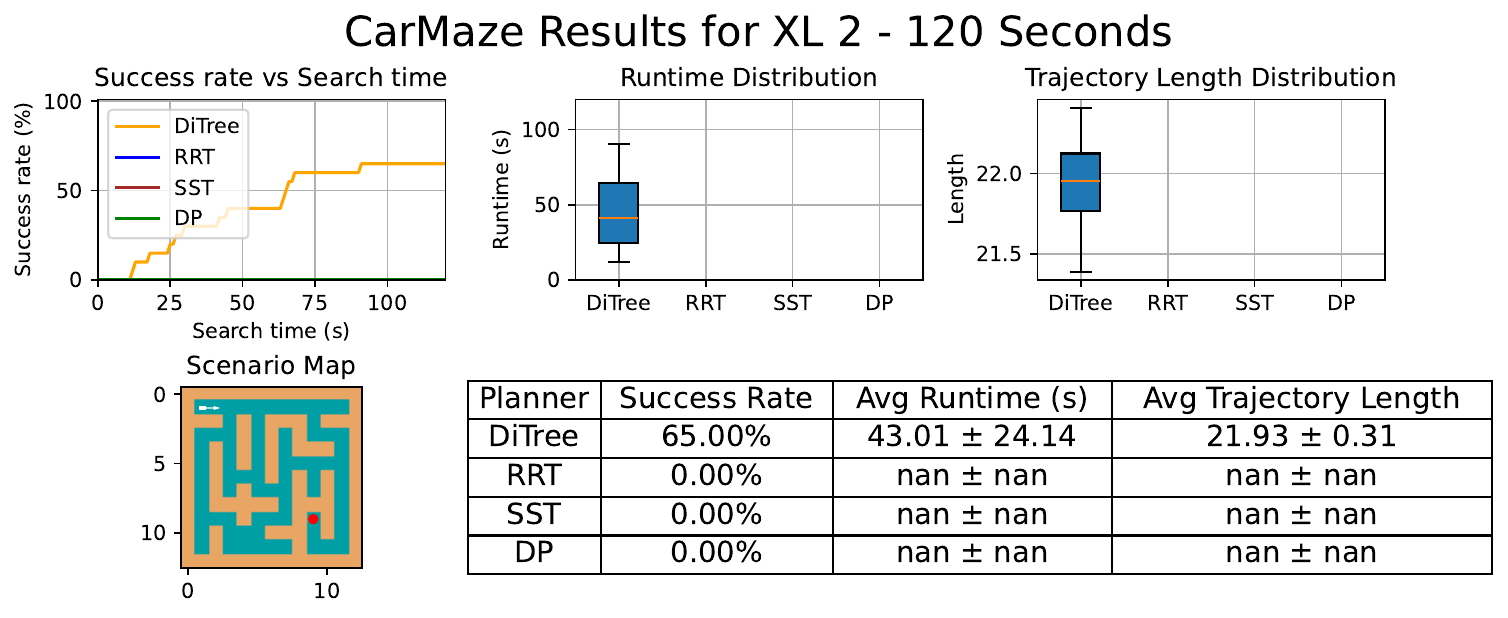}
\end{figure}

\begin{figure}[H]
    \centering
    \includegraphics[width=\textwidth]{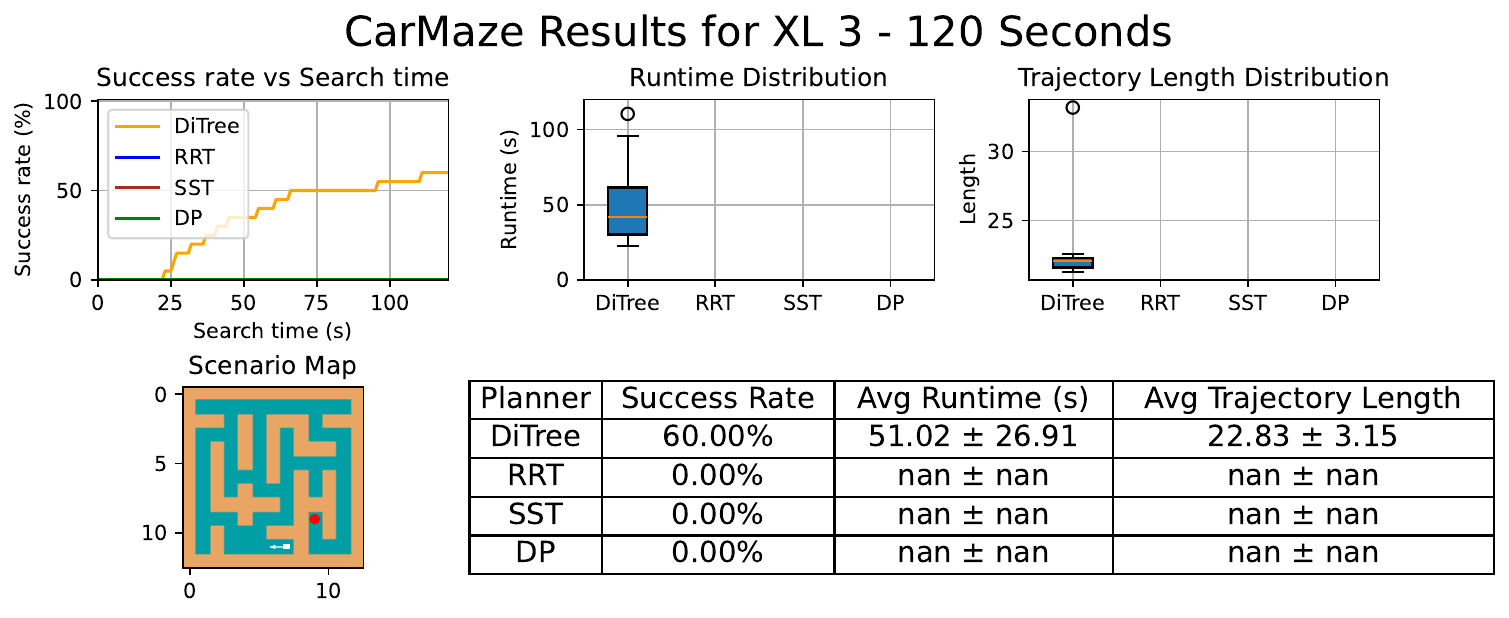}
\end{figure}

\begin{figure}[H]
    \centering
    \includegraphics[width=\textwidth]{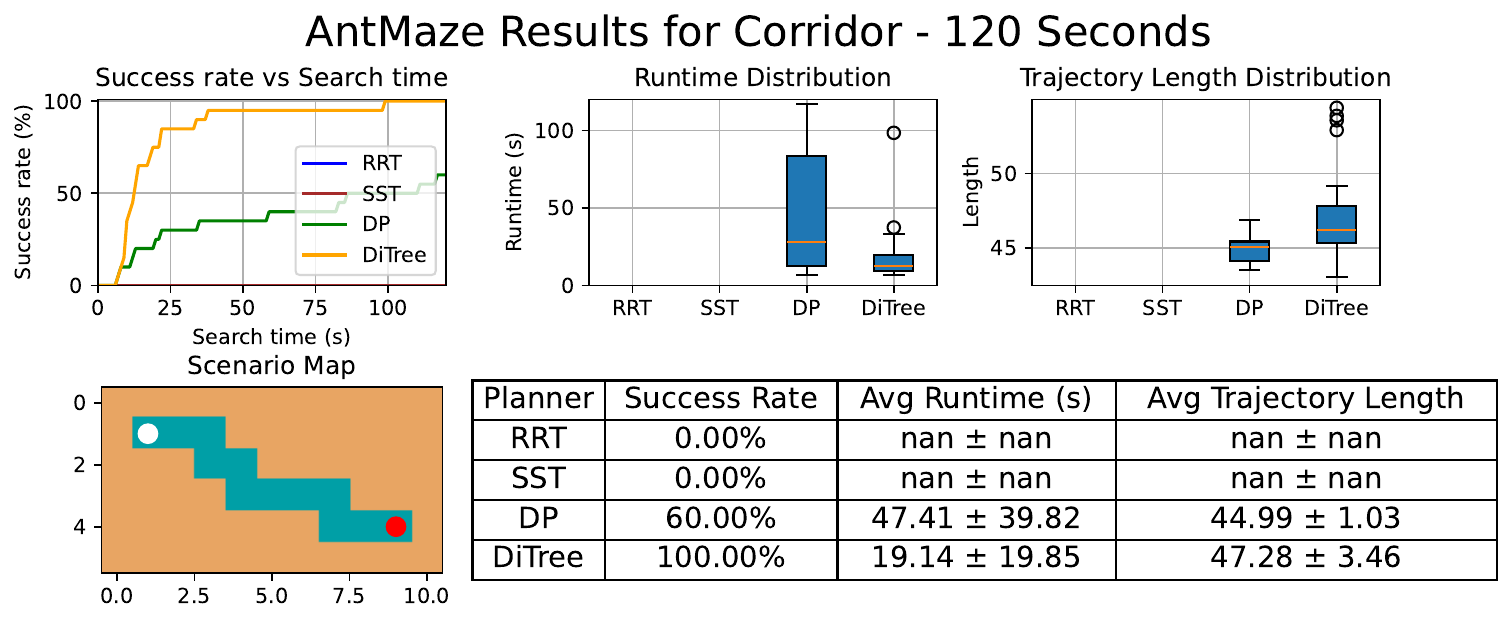}
\end{figure}

\begin{figure}[H]
    \centering
    \includegraphics[width=\textwidth]{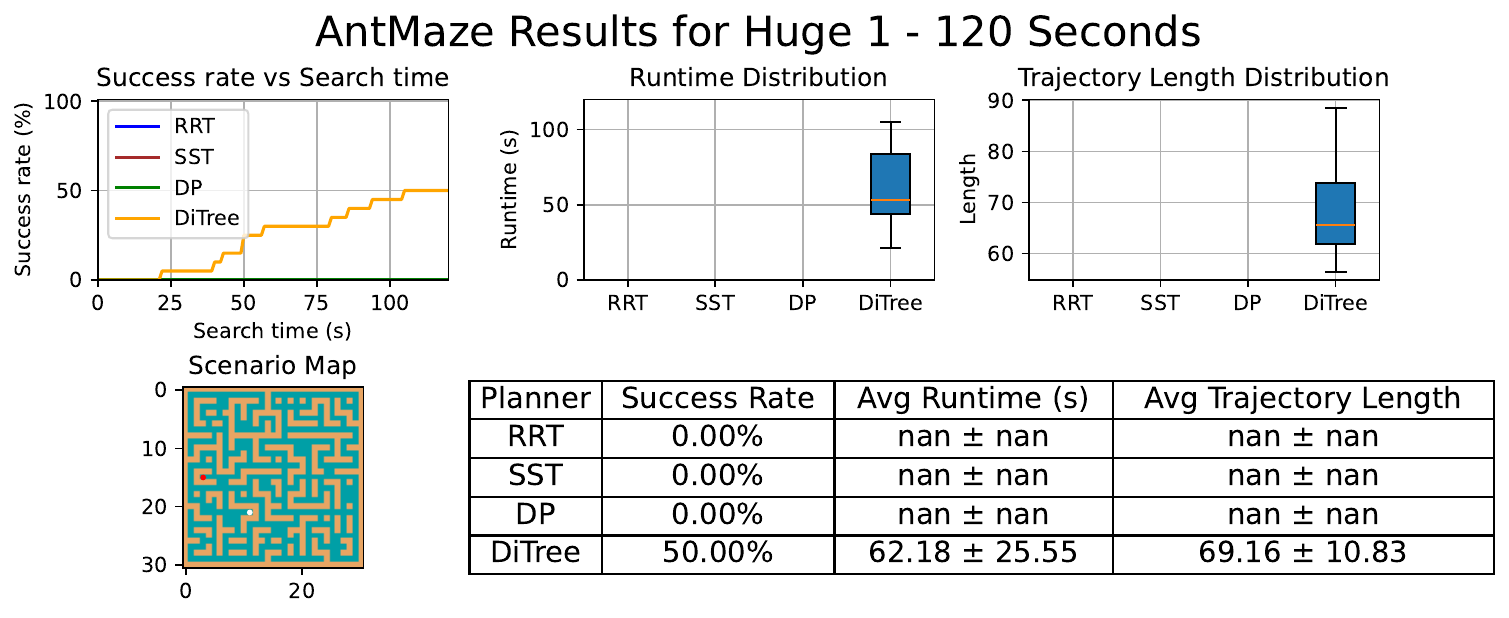}
\end{figure}

\begin{figure}[H]
    \centering
    \includegraphics[width=\textwidth]{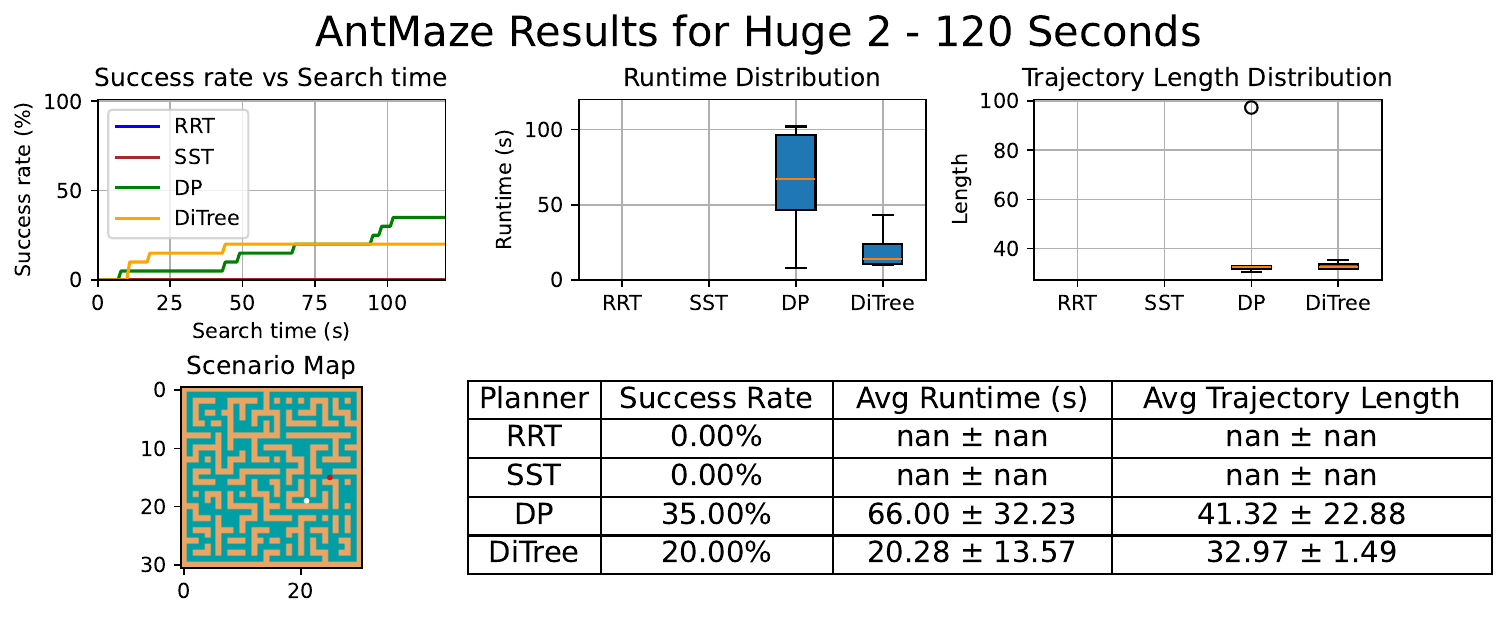}
\end{figure}

\begin{figure}[H]
    \centering
    \includegraphics[width=\textwidth]{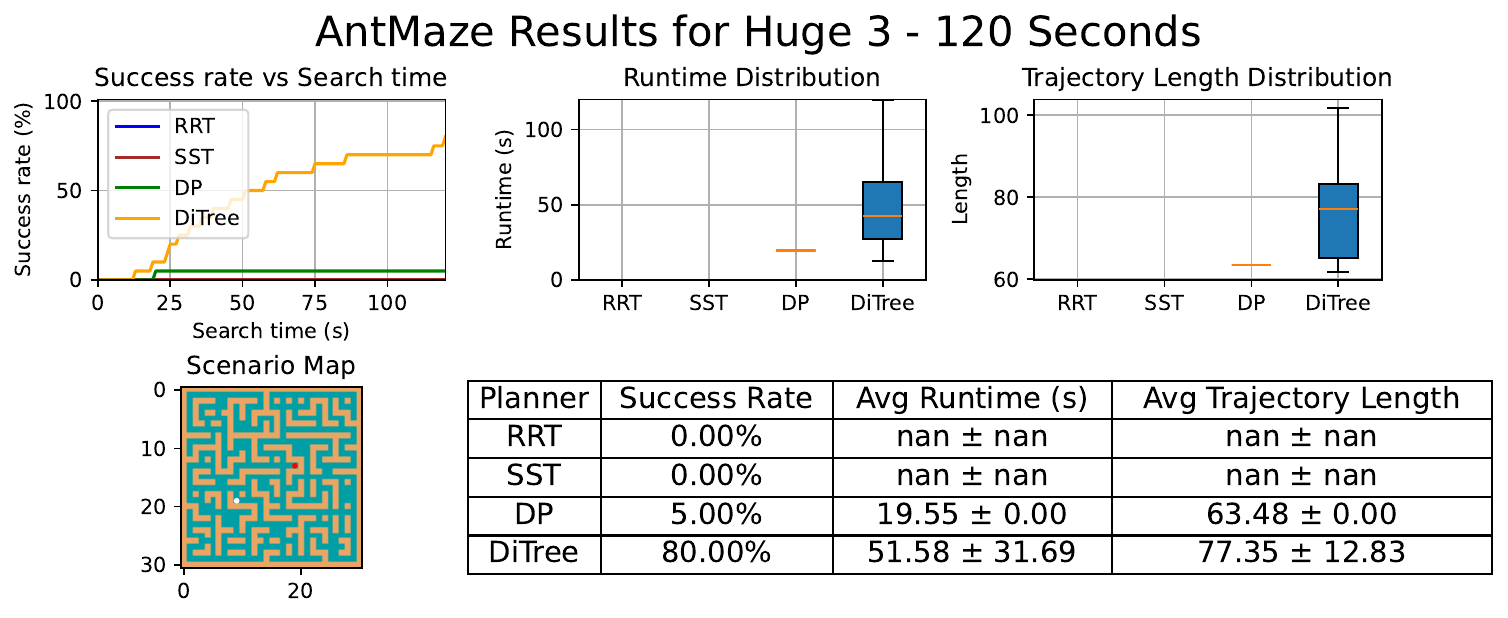}
\end{figure}

\begin{figure}[H]
    \centering
    \includegraphics[width=\textwidth]{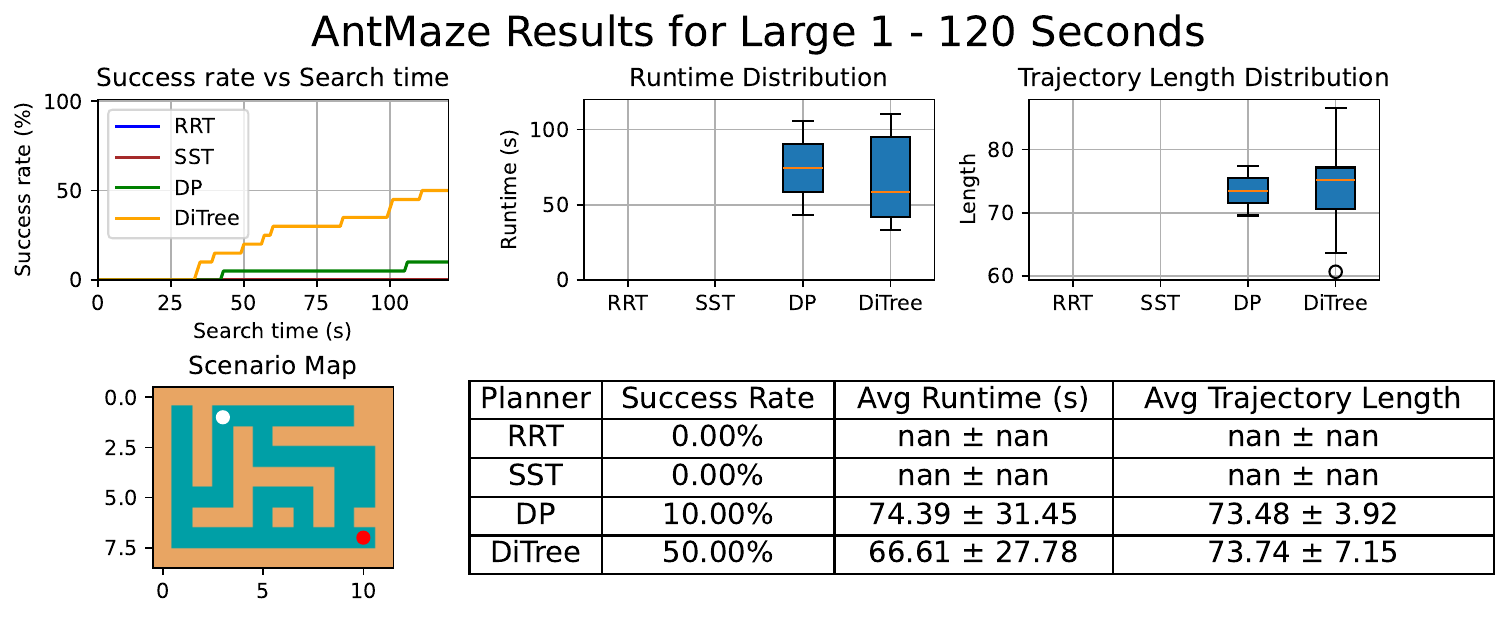}
\end{figure}

\begin{figure}[H]
    \centering
    \includegraphics[width=\textwidth]{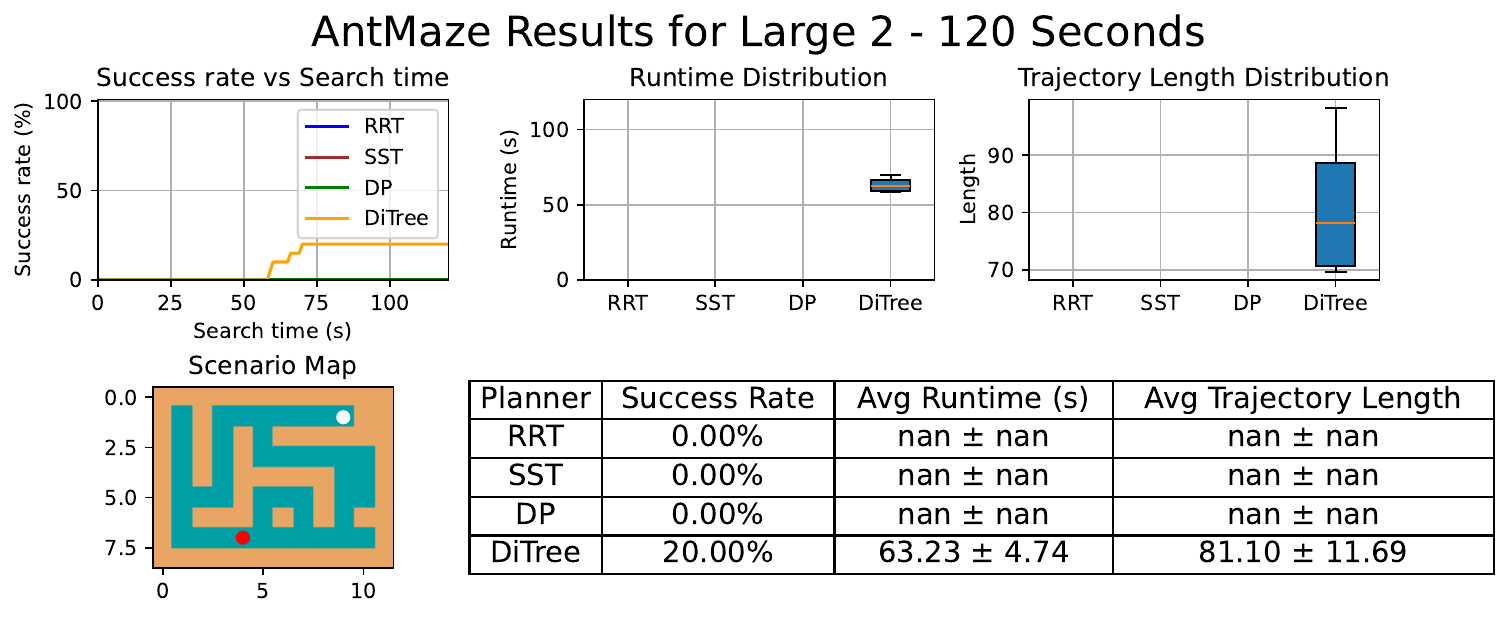}
\end{figure}

\begin{figure}[H]
    \centering
    \includegraphics[width=\textwidth]{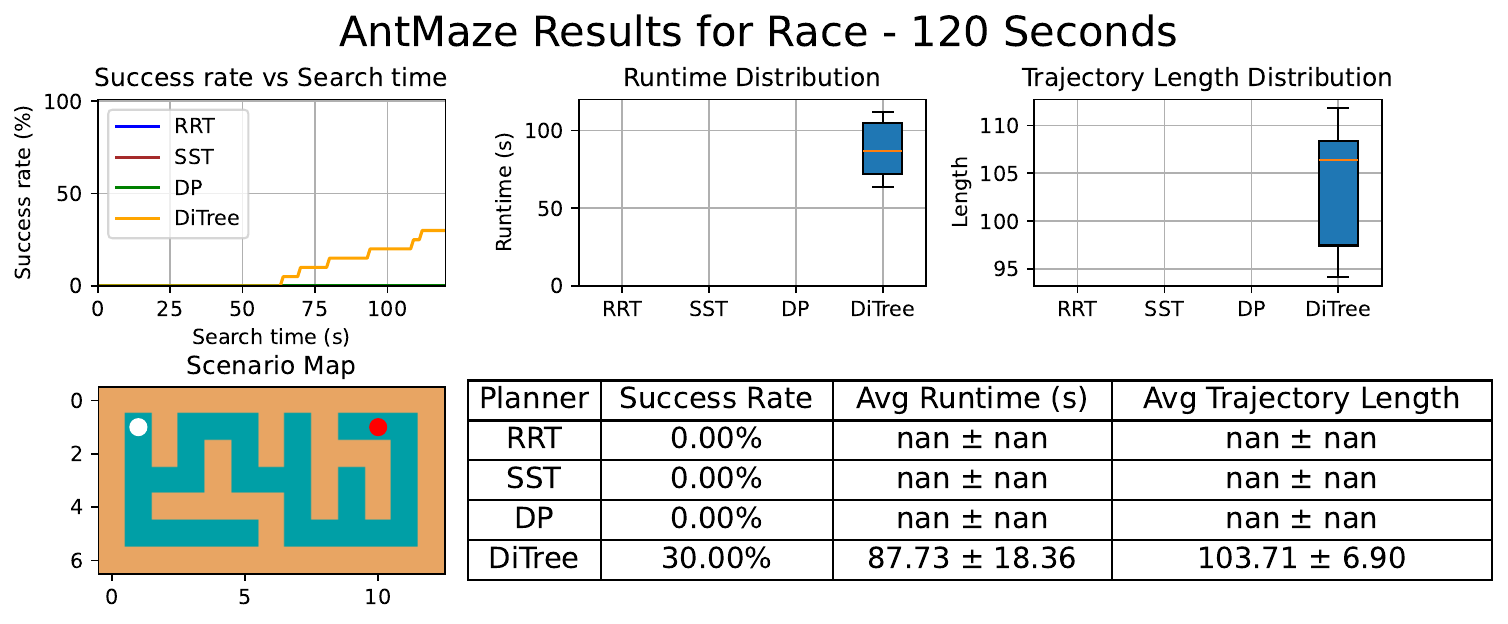}
\end{figure}

\begin{figure}[H]
    \centering
    \includegraphics[width=\textwidth]{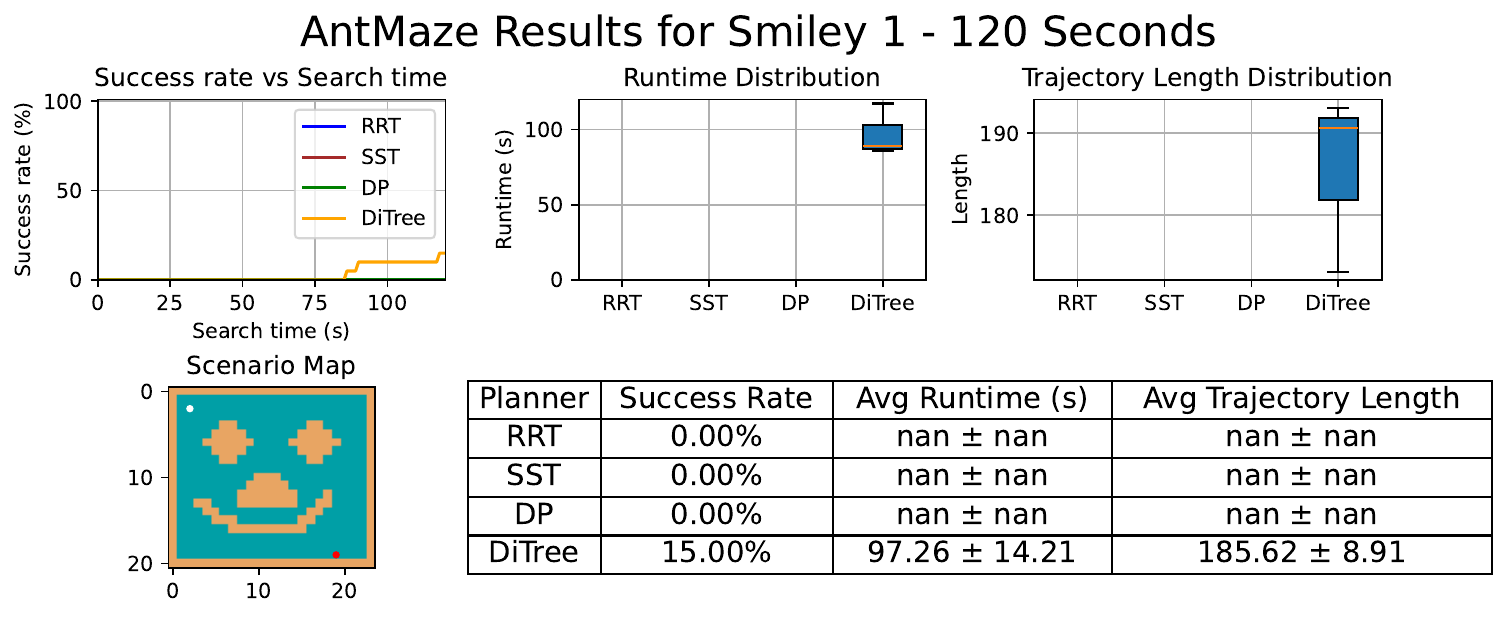}
\end{figure}

\begin{figure}[H]
    \centering
    \includegraphics[width=\textwidth]{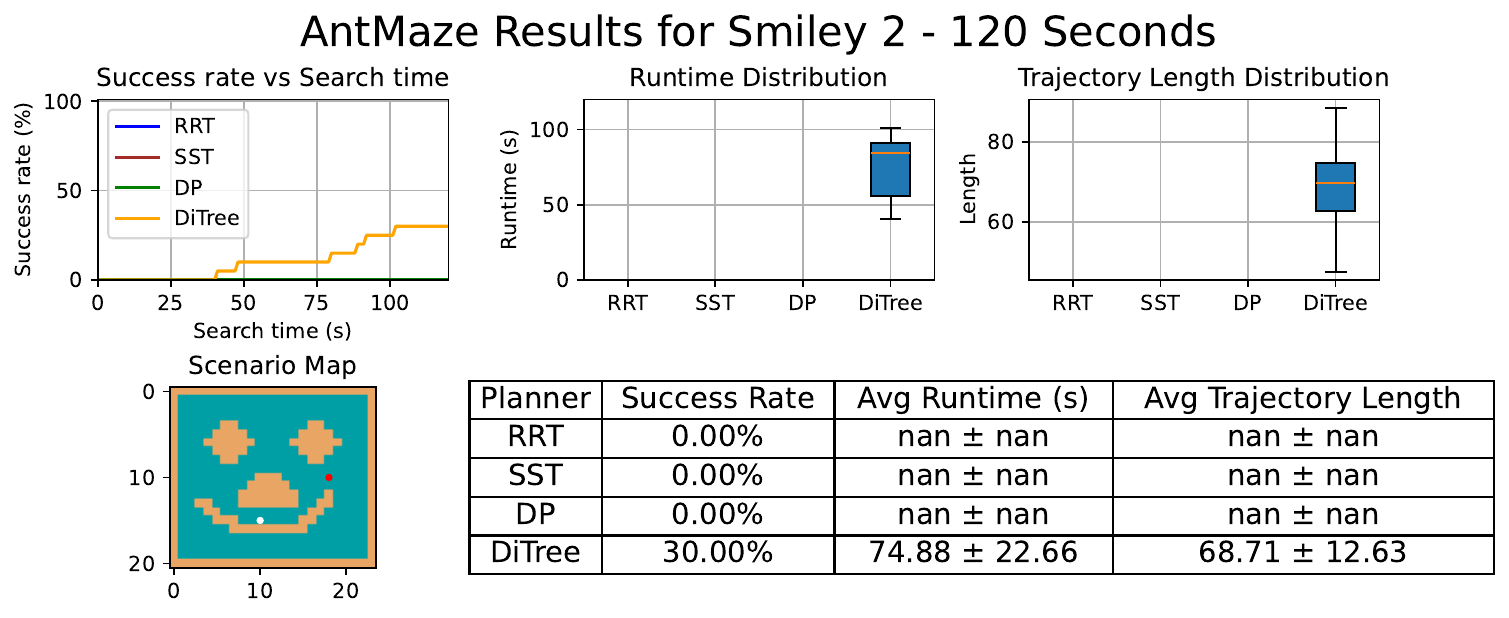}
\end{figure}

\begin{figure}[H]
    \centering
    \includegraphics[width=\textwidth]{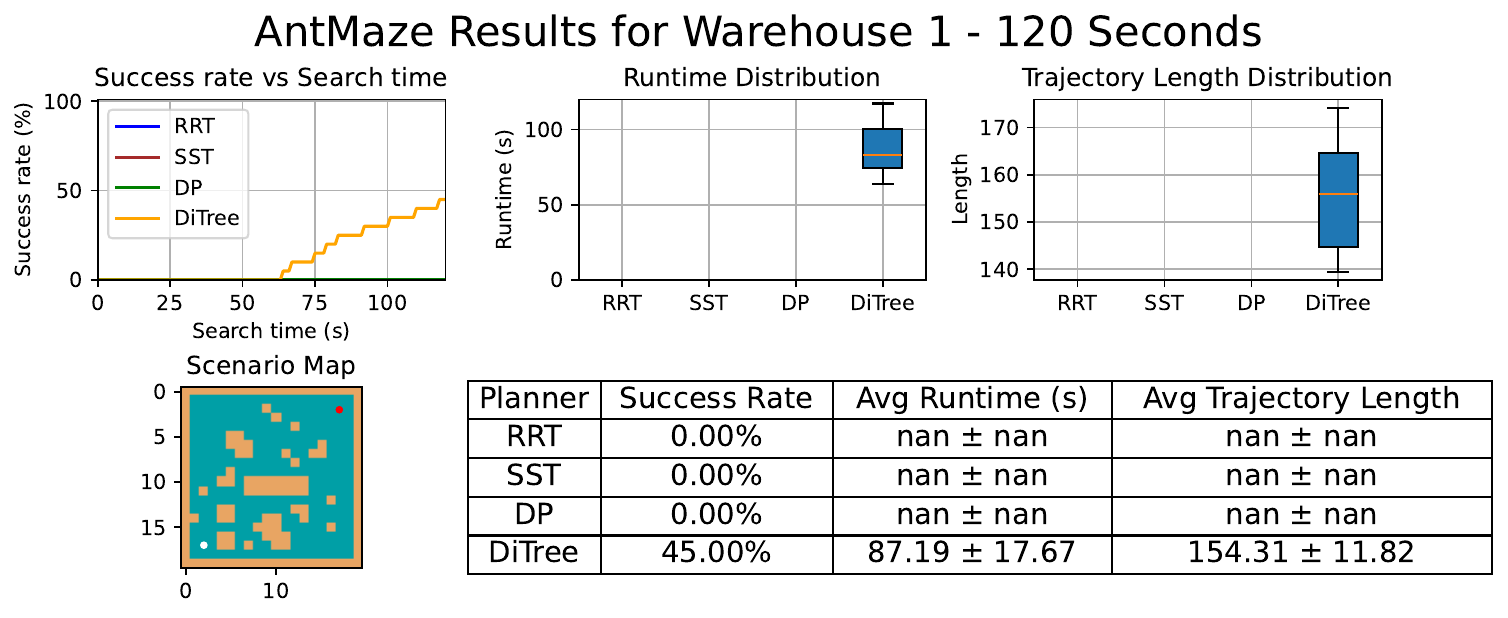}
\end{figure}

\begin{figure}[H]
    \centering
    \includegraphics[width=\textwidth]{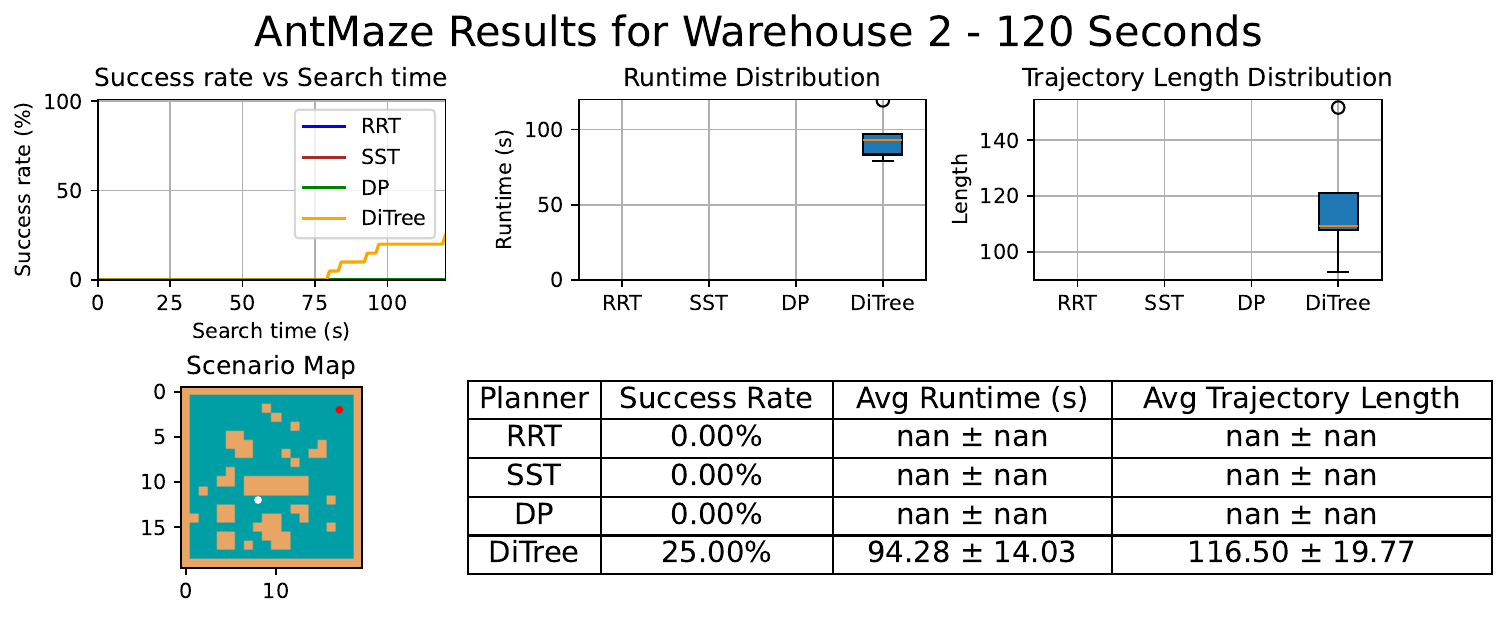}
\end{figure}

\begin{figure}[H]
    \centering
    \includegraphics[width=\textwidth]{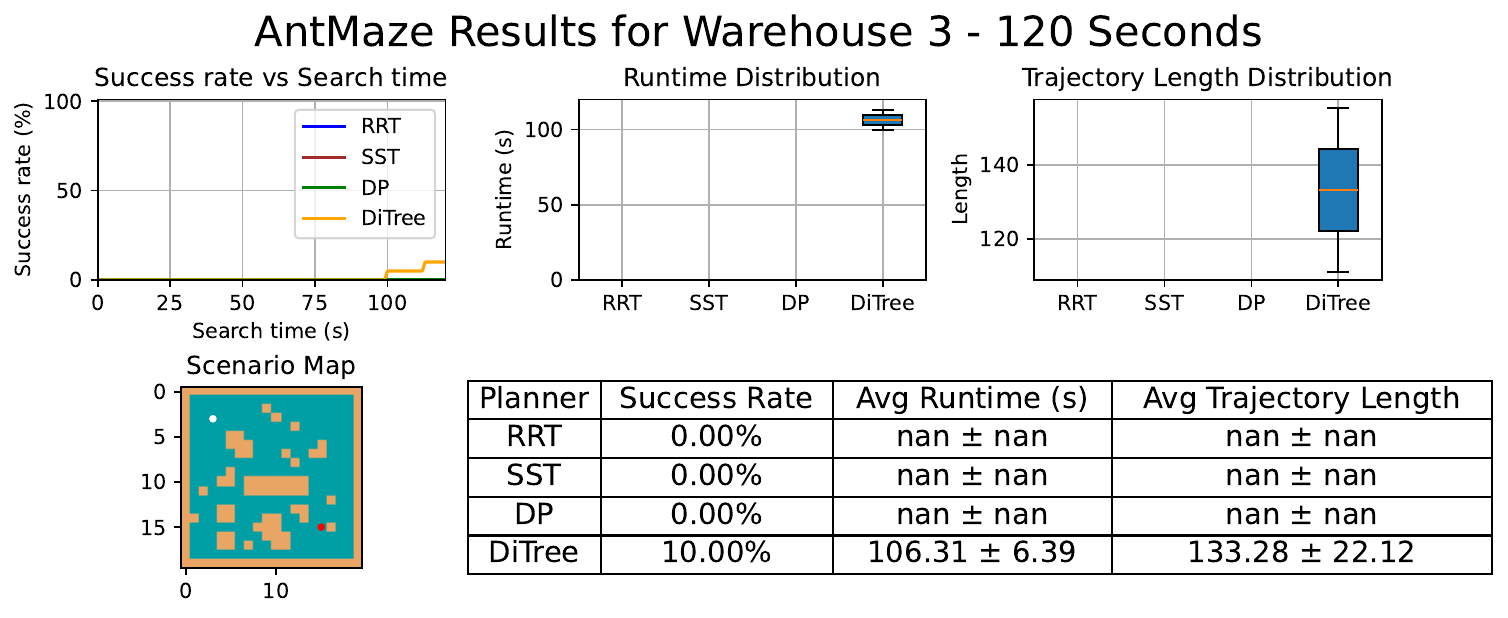}
\end{figure}

\begin{figure}[H]
    \centering
    \includegraphics[width=\textwidth]{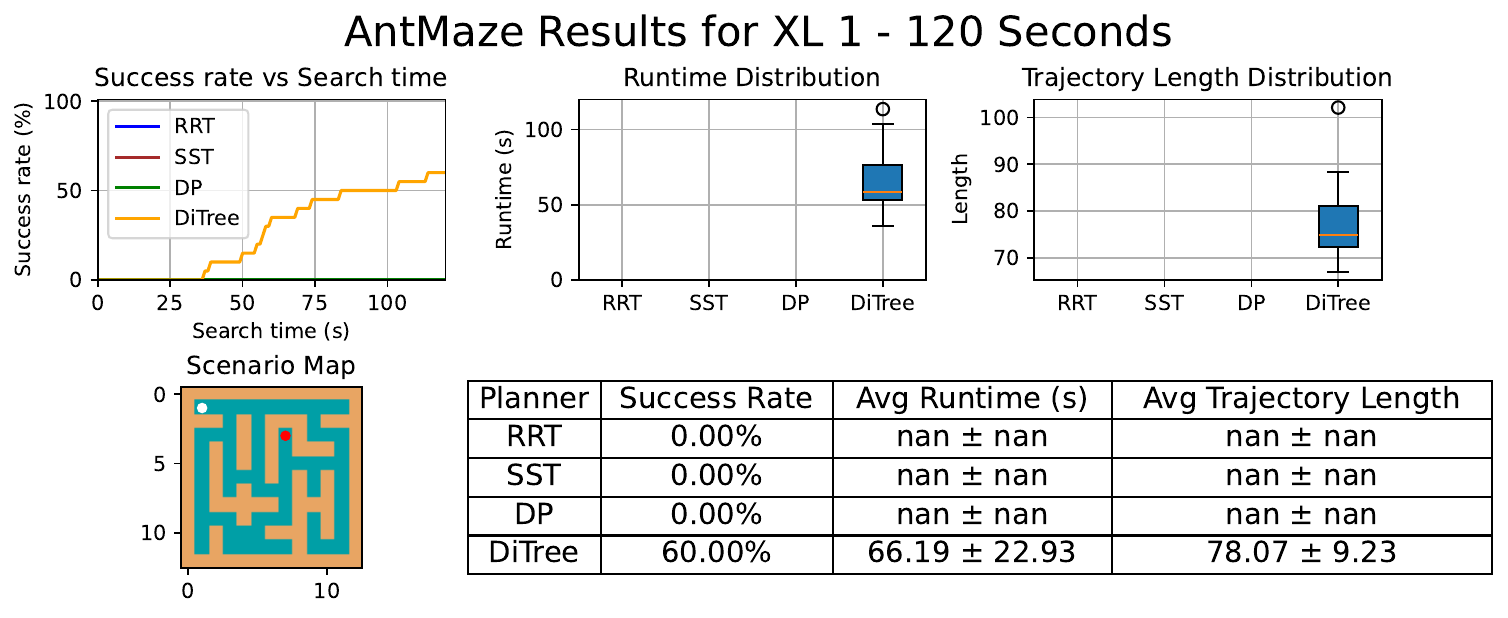}
\end{figure}

\begin{figure}[H]
    \centering
    \includegraphics[width=\textwidth]{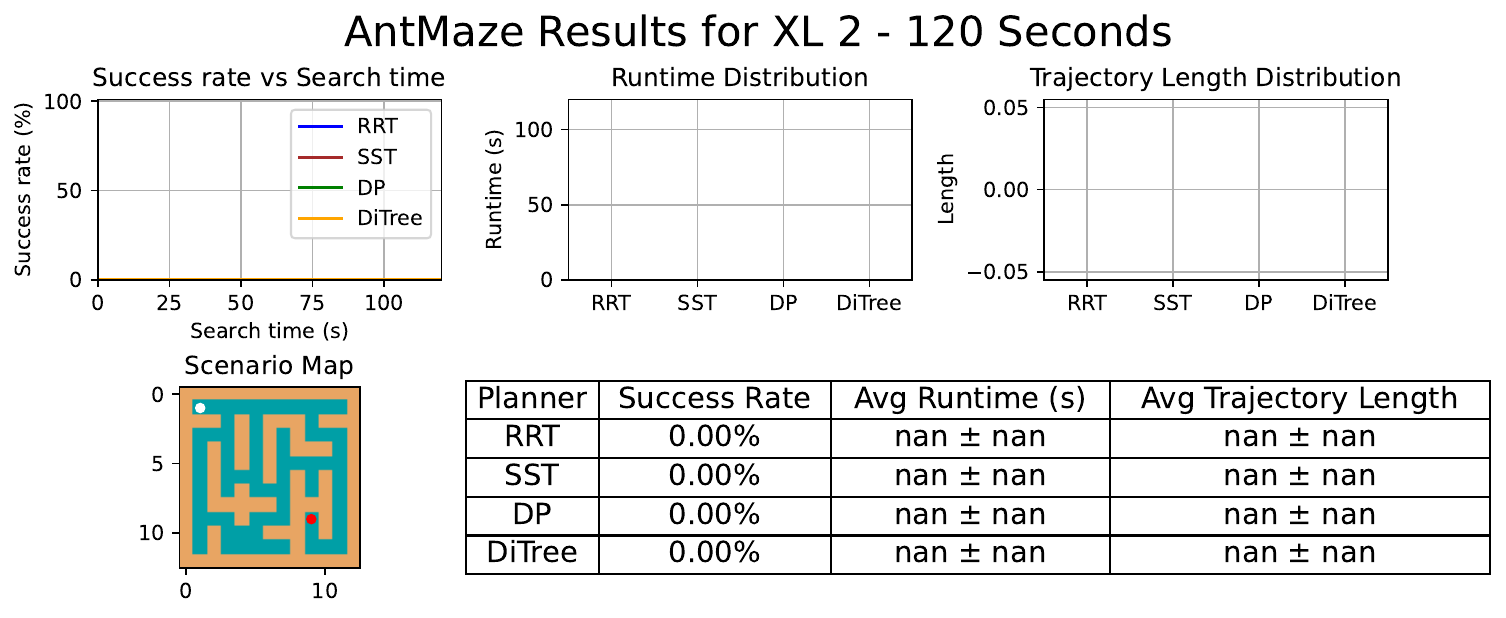}
\end{figure}

\begin{figure}[H]
    \centering
    \includegraphics[width=\textwidth]{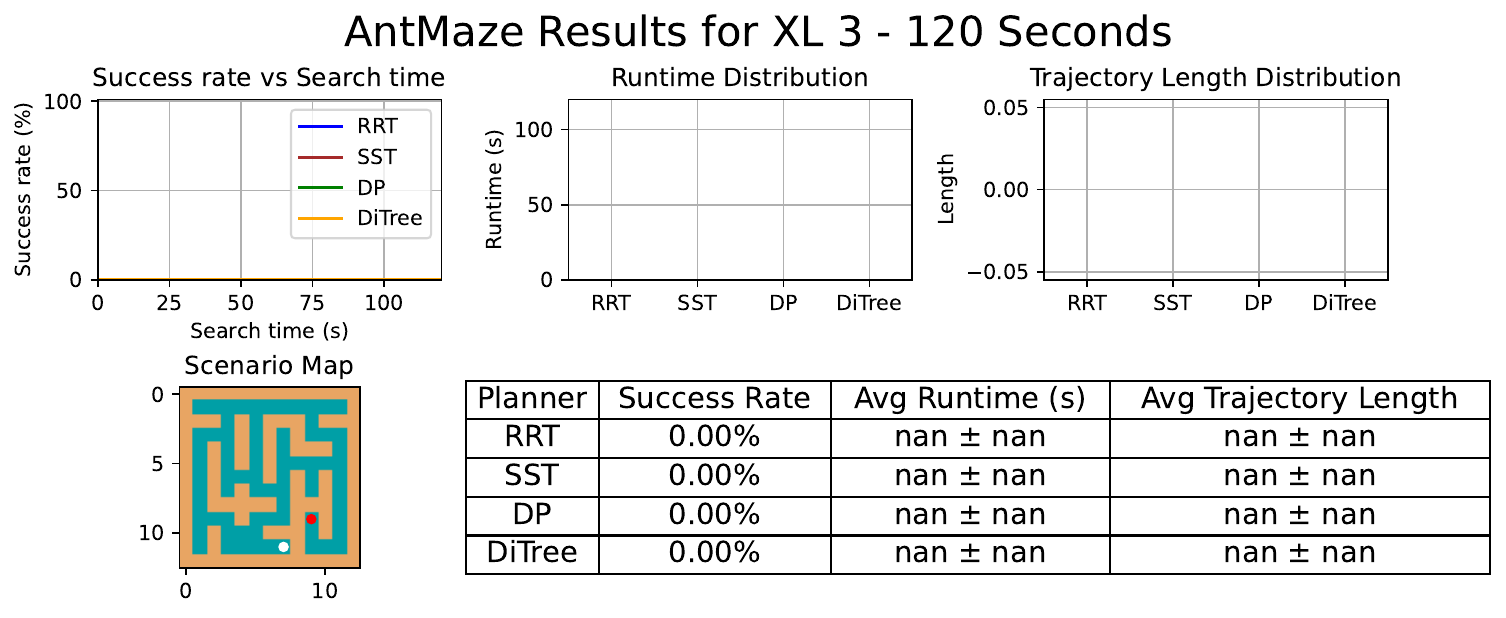}
\end{figure}

\end{document}